\documentclass[10pt,twocolumn,letterpaper]{article}

\usepackage{cvpr}              

\usepackage[accsupp]{axessibility}
\usepackage[utf8]{inputenc} 
\usepackage[T1]{fontenc}    
\usepackage{url}            
\usepackage{booktabs}       
\usepackage{amsfonts}       
\usepackage{nicefrac}       
\usepackage{microtype}      
\usepackage{caption}
\usepackage{graphicx}
\usepackage{amsmath}
\usepackage{amssymb}
\usepackage{booktabs}
\usepackage{threeparttablex}
\usepackage{multirow}
\usepackage{algorithm}
\usepackage{algpseudocode}
\usepackage{diagbox}
\usepackage{adjustbox}
\usepackage{footmisc}

\usepackage[dvipsnames, table]{xcolor}

\definecolor{cvprblue}{rgb}{0.21,0.49,0.74}
\usepackage[pagebackref,breaklinks,colorlinks,citecolor=cvprblue]{hyperref}

\def\paperID{3607} 
\def\confName{CVPR}
\def\confYear{2024}

\title{FreeMan: Towards Benchmarking 3D Human Pose Estimation \\ under Real-World Conditions}

\newcommand*{\affmark}[1][*]{\textsuperscript{#1}}
\newcommand*{\affaddr}[1]{#1} 

\author{%
Jiong Wang\affmark[1, 2]\footnotemark[1],~ 
Fengyu Yang\affmark[1]\footnotemark[1],
Bingliang Li\affmark[1],
Wenbo Gou\affmark[1], 
Danqi Yan\affmark[1],\\
Ailing Zeng\affmark[3],
Yijun Gao\affmark[2],
Junle Wang\affmark[2], 
Yanqing Jing\affmark[2],
Ruimao Zhang\affmark[1]\footnotemark[2]\\
\affaddr{\affmark[1]The Chinese University of Hong Kong, Shenzhen }
\affaddr{\affmark[2]Tencent }
\affaddr{\affmark[3]IDEA}\\
\small\url{jiongwang@link., fengyuyang1@link.cuhk.edu.cn, ruimao.zhang@ieee.org} \\
}

\begin{document}

\twocolumn[{
    \renewcommand\twocolumn[1][]{#1}
    \maketitle 
    \begin{center}
    \centering
    \includegraphics[width=\textwidth]{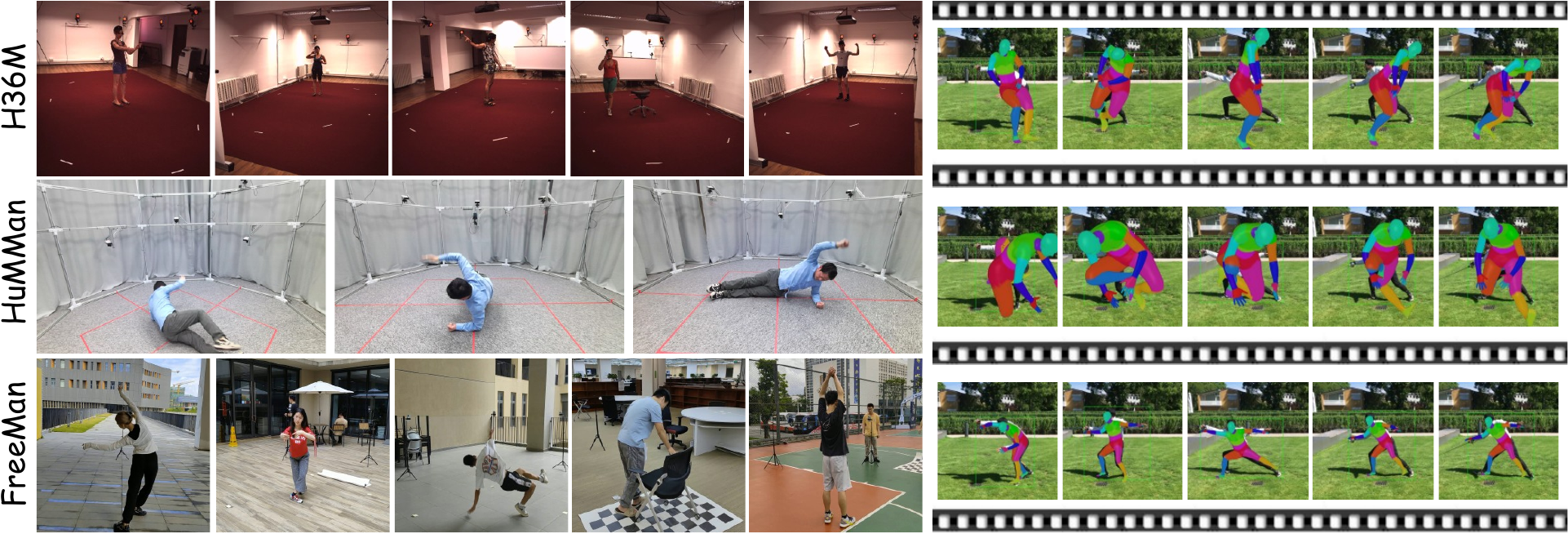}
    \captionof{figure}{
    The left displays sample frames from Human3.6M~\cite{h36m_pami} and HuMMan~\cite{cai2022humman}, which were collected under laboratory conditions, and contrasted with our FreeMan dataset that was collected in real-world scenarios. Frames from FreeMan have been cropped into a square format for visualization purposes, with the original resolution being $1920 \times 1080$ pixels. The right-hand side demonstrates the \textbf{test results on 3DPW of the HMR model}~\cite{hmrKanazawa17} trained on these three datasets. Notably, the model trained using FreeMan is able to adapt flawlessly to real-world conditions, demonstrating its superior generalization ability. Visualization uses implementation of mmHuman3D~\cite{mmhuman3d}.}
    \label{fig:datahighlight}
\end{center}

}]

\footnotetext[1]{First two authors contributes equally. Work done during Jiong Wang's MPhil study at CUHK(SZ). Wenbo Gou and Danqi Yan were research assistant at CUHK(SZ).}
\footnotetext[2]{Corresponding author. Email: ruimao.zhang@ieee.org}

\begin{abstract}
\vspace{-4mm}
Estimating the 3D structure of the human body from natural scenes is a fundamental aspect of visual perception. 
3D human pose estimation is a vital step in advancing fields like AIGC and human-robot interaction, serving as a crucial technique for understanding and interacting with human actions in real-world settings.
However, the current datasets, often collected under single laboratory conditions using complex motion capture equipment and unvarying backgrounds, are insufficient. The absence of datasets on variable conditions is stalling the progress of this crucial task. 
To facilitate the development of 3D pose estimation, we present FreeMan, the first large-scale, multi-view dataset collected under the real-world conditions. FreeMan was captured by synchronizing $8$ smartphones across diverse scenarios. It comprises $11M$ frames from $8000$ sequences, viewed from different perspectives. 
These sequences cover $40$ subjects across $10$ different scenarios, each with varying lighting conditions. 
We have also established an semi-automated pipeline containing error detection to reduce the workload of manual check and ensure precise annotation.
We provide comprehensive evaluation baselines for a range of tasks, underlining the significant challenges posed by FreeMan. 
Further evaluations of standard indoor/outdoor human sensing datasets reveal that FreeMan offers robust representation transferability in real and complex scenes. 
FreeMan is publicly available at \url{https://wangjiongw.github.io/freeman}.
\end{abstract}
\vspace{-3mm}
\makeatletter
\newcommand\figcaption{\def\@captype{figure}\caption}
\newcommand\tabcaption{\def\@captype{table}\caption}
\makeatother
\vspace{-2mm}
\section{Introduction}
\label{sec:intro}

Estimating 3D human poses from real scene input is a longstanding yet active research topic since its huge potential in real applications, such as animation creation~\cite{Weng_2022_CVPR_humannerf,yoon2021poseanimation}, virtual reality~\cite{thomas2018virtualbody,grover2021pipeline}, the metaverse~\cite{lee2021allmetaverse,yoon2022metaverse,ljungholm2022metaverse} and human-robot interaction~\cite{hentout2019human}.
Specifically, it aims to identify and determine the spatial positions and orientations of the human body's parts in 3D space from input data such as the image or the video. 
Despite numerous models proposed in recent years~\cite{DBLP:journals/corr/abs-2104-08527,DBLP:journals/corr/abs-2111-12707,wang2021mvp}, practical implementation in real scenes remains challenging due to the varying conditions such as viewpoint, occasions, human scale, uneven light conditions and complex background.
Some challenges may stem from the disparity between the recent benchmarks and real-world scenarios.
As shown in Fig.~\ref{fig:datahighlight}, the widely recognized Human3.6M~\cite{h36m_pami}, along with the currently largest dataset HuMMan~\cite{cai2022humman}, 
are usually in laboratory settings utilizing intricate equipment, which maintains constant camera parameters and offers minimal variation in background conditions.
%
%
The effectiveness of the trained models when trained using these datasets often decline significantly in real-world environments.
%
%

From a data-oriented perspective, we have identified several constraints that hinder the performance of the existing models.
\textbf{(1) Insufficient Scene Diversity.}
Existing datasets, as shown in Tab.~\ref{tab:OverviewDatasets}, are mainly collected in controlled laboratory conditions, which may not be optimal for robust model training due to static lighting conditions and uniform backgrounds. This limitation becomes especially crucial when the objective is to estimate 3D pose in real-world scenarios, where scene contexts exhibit substantial variability.
%
%
In certain datasets, even though the data is collected from outdoor scenes, \textit{e.g.,} MuCo~\cite{muco_singleshotmultiperson2018} and 3DPW~\cite{vonMarcard2018} in Tab.~\ref{tab:OverviewDatasets}, the variety of scenarios remains remarkably limited. 
This constraint significantly hampers the applicability of trained models across a broader range of situations.
\textbf{(2) Limited Actions and Body Scales.} In existing datasets, the range of human actions tends to be rather limited. Even in the currently largest dataset, HuMMan~\cite{cai2022humman}, the variety of actions in the publicly available data is quite restricted. Additionally, these large datasets typically employ fixed cameras to capture data from various perspectives. The distance from the camera to the actor is relatively constant, which results in a relatively fixed human body scale across different videos.
%
%
%
%
\textbf{(3) Restricted Scalability.} The annotation of current datasets primarily relies on expensive manual processing, which greatly restricts the scalability of the datasets. Especially when the camera used for collection is movable, how to effectively align data from different cameras and perform efficient annotation remains an open issue.

\begin{table}
    
    \begin{threeparttable}
    \begin{adjustbox}{max width=\linewidth}
    
    \begin{tabular}{l|c|c|c|c|c|c|c|c}
    \hline
    Dataset & Environment & \#Subj & \#Action & \#Scene & \#Seq & \#Frame & \#Camera & FPS \\
    \hline
    HumanEva\cite{humaneva/ijcv/SigalBB10} & Laboratory & 4 & 6 & 1 & 168 & 80K & 7 & 30 \\
    CMU Panoptic\cite{Joo_2017_CMU_TPAMI} & Laboratory & 8 & 5 & 1 & 65 & 154M & 31 & 30 \\
    MPI-INF-3DHP\cite{mono-3dhp2017} & Real Scene & 8 & 8 & 1 & 16 & 1.3M & 14 & 30 \\
    3DPW\cite{vonMarcard2018} & Real Scene & 7 & 47 & 4 & 60 & 51K & 1 & 30 \\
    Mirrored Human\cite{fang2021mirrored} & Laboratory & - & - & - & - & 1.5M & 1 & 30 \\
    \rowcolor{gray!15} Human3.6M\cite{h36m_pami} & Laboratory & 9 & 15 & 1 & 840 & 3.6M & 4 (Fixed) & 30 \\
    \rowcolor{gray!15} AIST++\cite{li2021aistpp} & Laboratory & 30 & 10 & 1 & 1408 & 10.1M & 9 (Fixed) & 30 \\
    \rowcolor{gray!15} HuMMan\cite{cai2022humman} & Laboratory & 1000 & 500 & 1 & 400K & 60M\textcolor{red}{\dag} & 11 (Fixed) & 30 \\
    \rowcolor{gray!15} HuMMan-released\cite{cai2022humman} & Laboratory & 132 & 20 & 1 & 4466 & 278K\textcolor{red}{\dag} & 11 (Fixed) & 30 \\
    \rowcolor{gray!15} FreeMan & Real Scene & 40 & 123 & 10\textcolor{blue}{\ddag} & 8000 & 11.3M & 8 (Movable) & 30 / 60 \\
    \hline
    \end{tabular}
    \end{adjustbox}
    
    \caption{Overview of 3D human pose datasets. $^{1}$ Comparison of our proposed FreeMan dataset with existing 3D Human Pose datasets. Only HD Cameras counted for CMU Panoptic\cite{Joo_2017_CMU_TPAMI}. $^{\textcolor{red}{\dag}}$ Only 1\% of the HuMMan dataset (600K frames) is made publicly available. $^{\textcolor{blue}{\ddag}}$ FreeMan includes 10 types of scenes that correspond to 29 locations. Fixed means cameras are fixed within the whole dataset, while our cameras are movable and camera poses vary among video sequences.}
    \label{tab:OverviewDatasets}
    \end{threeparttable}
\end{table}

\begin{figure}
    \centering
    \includegraphics[width=\linewidth]{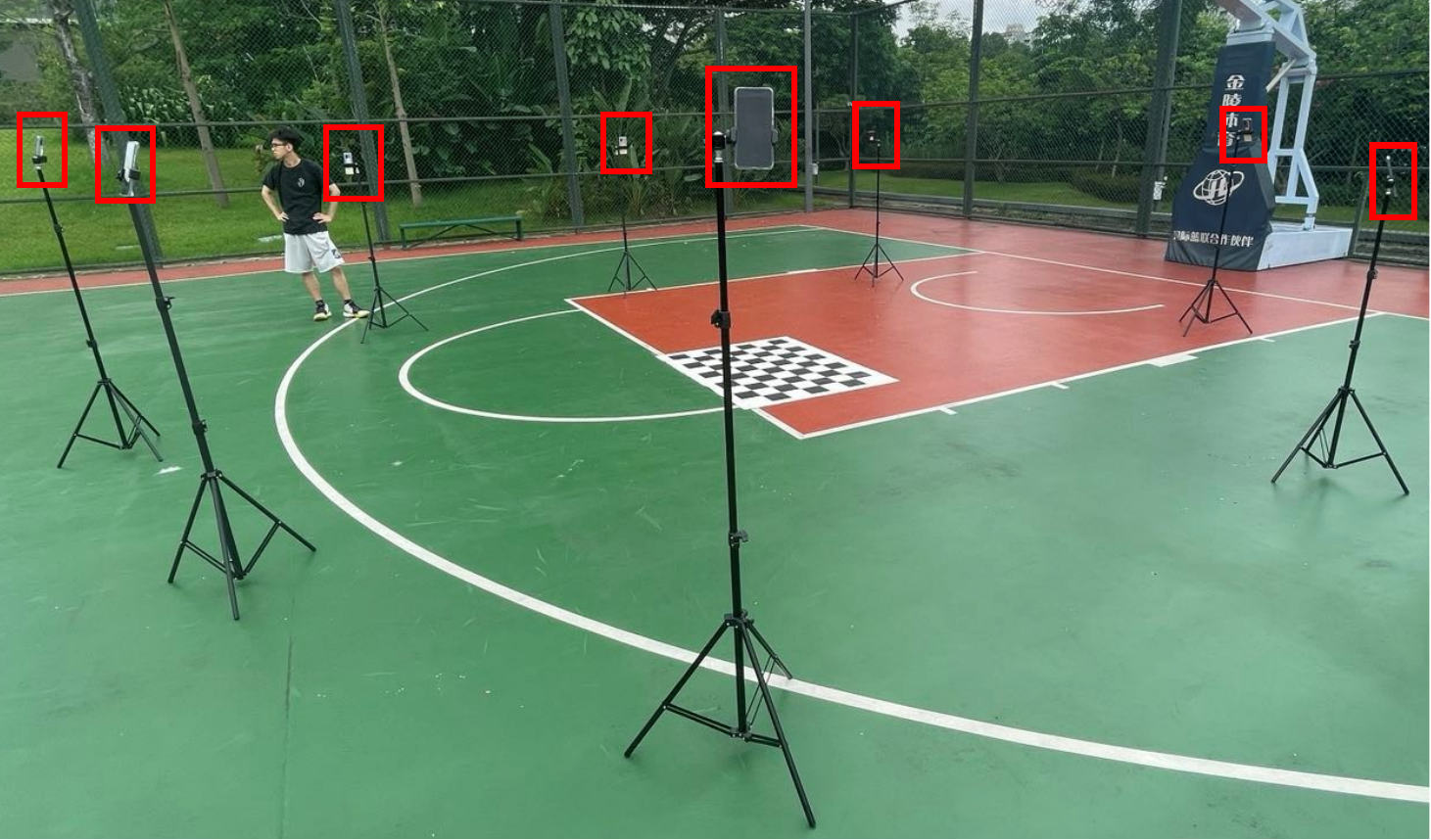}
    \caption{Equipment setting of data collection using 8 cameras. Cameras are attached to tripods.}
    \label{fig:device_settings}
    \vspace{-3mm}
\end{figure}

To address these above issues, this work presents FreeMan, a novel large-scale benchmark for 3D human pose estimation under real-world conditions.
FreeMan contains $11M$ frames in $8000$ sequences captured by $8$ smartphone cameras from different views simultaneously, as illustrated in Fig.~\ref{fig:device_settings}.
It covers $40$ subjects in $10$ kinds of scenes. 
To our best knowledge, it is the current largest multi-view 3D pose estimation dataset, with variable camera parameters and complex background environments. 
It is $215\times$ of the famous outdoor dataset 3DPW~\cite{vonMarcard2018}.  
From a practical perspective, it has several appealing strengths:
\textbf{Firstly}, a large number of scenes introduce diversity in both backgrounds and lighting, enhancing the generalization ability of models trained on FreeMan in real-world scenarios. This makes it particularly suitable for evaluating algorithmic performance in practical applications.
\textbf{Secondly}, the distances between the $8$ cameras and the actors are variant (\textit{i.e.,} $2$ to $5.5$ meters) across sequences, resulting in significant scale changes in human bodies.
\textbf{Thirdly}, although we employed mobile RGB cameras to collect data, we propose a semi-automated annotation pipeline and erroneous frame detection, thereby significantly reduce manual workload and enhance the scalability and annotation accuracy of the dataset.
\textbf{Lastly}, the proposed FreeMan encompasses a wide range of pose estimation tasks, which include monocular 3D estimation, 2D-to-3D lifting, multi-view 3D estimation, and neural rendering of human subjects.
We present thorough evaluation baselines for the aforementioned tasks on FreeMan, highlighting the inherent challenges of such a new benchmark.


In summary, this paper has made three contributions:
\begin{itemize}
    \item We have constructed a large-scale dataset for 3D human pose estimation under varied real-world conditions. 
The impressive transferability of the models trained on FreeMan to real-world scenarios has been demonstrated.
\item  We have showcased a simple yet effective toolchain that enables the semi-automatic annotation and efficient manual correction.
%
\item  We provide comprehensive benchmarks for human pose estimation and modeling on FreeMan, facilitating downstream applications. These baselines highlight potential directions for future algorithmic enhancements. 
\end{itemize}

\section{Related Work}
\label{sec:relwork}

\textbf{Human Pose Datasets.} 
Human modeling is a significant task in computer vision. Existing datasets predominantly rely on 2D and 3D keypoint annotations, with 3D keypoint datasets available in two forms: monocular and multi-view.
For 2D keypoint, there are some single-frame datasets such as MPII~\cite{andriluka14cvpr} and COCO~\cite{cocodataset}, which provide diverse images with 2D keypoints annotations, while video datasets such as J-HMDB~\cite{Jhuang:ICCV:2013}, Penn Action~\cite{6751390} and PoseTrack~\cite{DBLP:journals/corr/abs-1710-10000} provide 2D keypoints with temporal information. 
In contrast, 3D keypoint datasets are often constructed in indoor scenes, such as Human3.6M~\cite{h36m_pami}, CMU Panoptic~\cite{Joo_2015_ICCV}, MPI-INF-3DHP~\cite{mono-3dhp2017}, AIST++~\cite{li2021learn}  and HuMMan~\cite{cai2022humman} for multi-view. There also exists some outdoor datasets such as 3DPW~\cite{vonMarcard2018} for monocular cases. Details of these datasets are shown in Tab.~\ref{tab:OverviewDatasets}. However, the majority of outdoor datasets such as MPI-INF-3DHP, MuCo-3DHP, and 3DPW exhibit a limited variety of acquisition scenes, and the datasets that involve fixed camera poses such as AIST++.

\noindent\textbf{3D Human Pose Estimation.} 
The present study categorizes the task of 3D pose estimation into three distinct types, namely 2D-to-3D pose lifting, monocular 3D pose estimation, and multi-view 3D pose estimation. In the 2D-to-3D pose lifting task, Martinez~\cite{DBLP:journals/corr/MartinezHRL17} proposed a simple baseline to regress the 3d keypoints based on a convolutional neural network from 2D keypoints. However, subsequent works, such as Videopose3D~\cite{DBLP:journals/corr/abs-1811-11742}, PoseFormer~\cite{DBLP:journals/corr/abs-2103-10455} and MHFormer~\cite{DBLP:journals/corr/abs-2111-12707}, have improved upon this baseline by integrating temporal information into their models. In monocular 3D pose estimation task, HMR~\cite{hmrKanazawa17}, SPIN~\cite{kolotouros2019spin} takes a single RGB image as input to perform 3D huna pose estimation, which is often used as baselines for comparison with other algorithms, such as PARE~\cite{DBLP:journals/corr/abs-2104-08527}, SPEC~\cite{SPEC} and HybrIK~\cite{Hybrik}. Additionally, multi-view methods are proposed to accommodate potential body parts overlapping in monocular view. Iqbal's~\cite{iqbal2020weakly} and MCSS~\cite{mitra2020multiview} adopt weak supervision to reduce the dependence on the 3D annotated pose, while Canonpose~\cite{wandt2021canonpose} and EpipolarPose~\cite{kocabas2019self} turned to self-supervise fashion to deal with multi-view data.

\noindent\textbf{Neural Rendering of Human Subjects.}
With the development of NeRF~\cite{mildenhall2021nerf} in dynamic scene rendering, people also focus on the dynamic rendering of humans. Compared to dynamic scenes, the non-rigid property of humans has more challenges. The prior knowledge of body movements can provide a good prior for rendering, and many methods use SMPL~\cite{SMPL:2015} as a prior for body rendering. Most methods reconstruct human bodies through multi-view videos~\cite{liu2020neural,noguchi2021neural,weng2020vid2actor}, while recent works have also employed single-view videos, such as HumanNeRF~\cite{weng2022humannerf}, FlexNeRF~\cite{jayasundara2023flexnerf}, YOTO~\cite{kim2023you}. 

\section{FreeMan Dataset}
\label{sec:datasets}

FreeMan is a large-scale multi-view dataset under real-world conditions with precise 3D pose annotations. It comprises $11M$ frames from $1000$ sessions, featuring $40$ subjects across $10$ types of scenes. The dataset includes $10M$ frames recorded at $30$FPS and an additional $1M$ frames at $60$FPS.
Next, we highlight the diversity of FreeMan, from various scenario selections, actions, camera settings and subjects.

\begin{figure}[t]
    \centering
    \includegraphics[width=\linewidth]{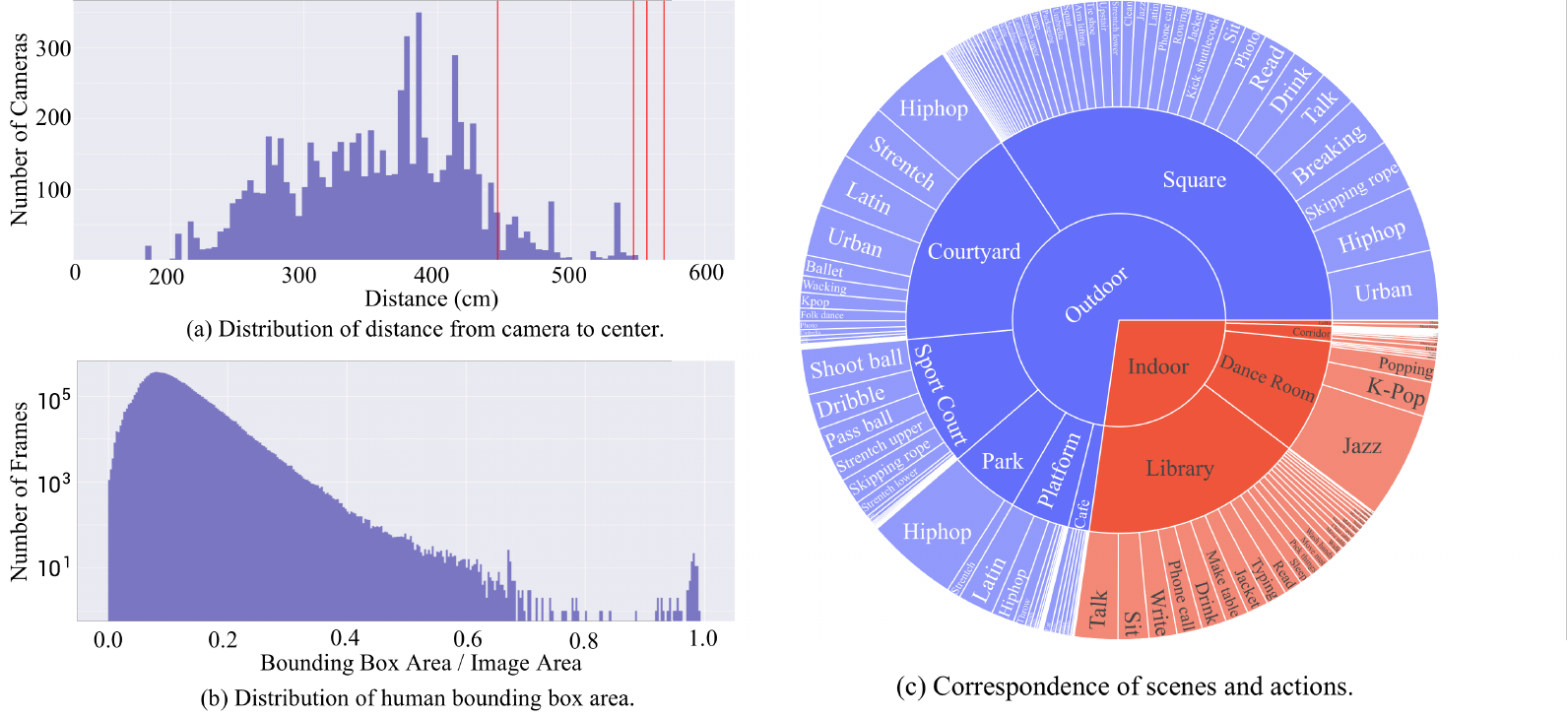}
    \captionsetup{font={small}}
    \caption{(a) Distribution of distance from the camera to the center of the system, indicated by translation along the z-axis in camera parameters. Four vertical red lines represent the distance of 4 cameras in Human3.6M~\cite{h36m_pami}. (b) Distribution of human bounding box areas. The horizontal axis represents the ratio of the bounding box area over the image area. The vertical axis is in log scale. (c) Correspondence of scenes and actions. Areas of blocks represent the scale of the respective frame number. The outmost circle shows actions and the circle in the middle present $10$ type of scenes in our dataset. Zoom in $10 \times$ for the best view.}
    \label{fig:dataset_stat}
    \vspace{-3mm}
\end{figure}

\noindent\textbf{Scenarios.}
We design $10$ types of real-world scenes, including $4$ indoor and $6$ outdoor scenes, for our data collection. Fig.~\ref{fig:dataset_stat} (c) illustrates the scene diversity of our FreeMan.
The blue section represents the outdoor part, while the red part refers to frames captured in indoor scenes. 
Specifically, there are $2.76$ million frames captured indoors and $8.45$ million frames captured outdoors. 
In the outdoor data, there are different frame numbers collected under varying lighting conditions, with $1$ million frames captured at night or dusk and $7.45$ million frames captured during daytime. 
Moreover, the central block of the circle denotes different scenarios, while the blocks on the outermost circle refer to actions. 
The areas of the blocks are proportional to frame number. 
Please refer to supplementary material for more details.
%

\begin{figure*}[!t]
    \centering
    \includegraphics[width=0.9\linewidth]{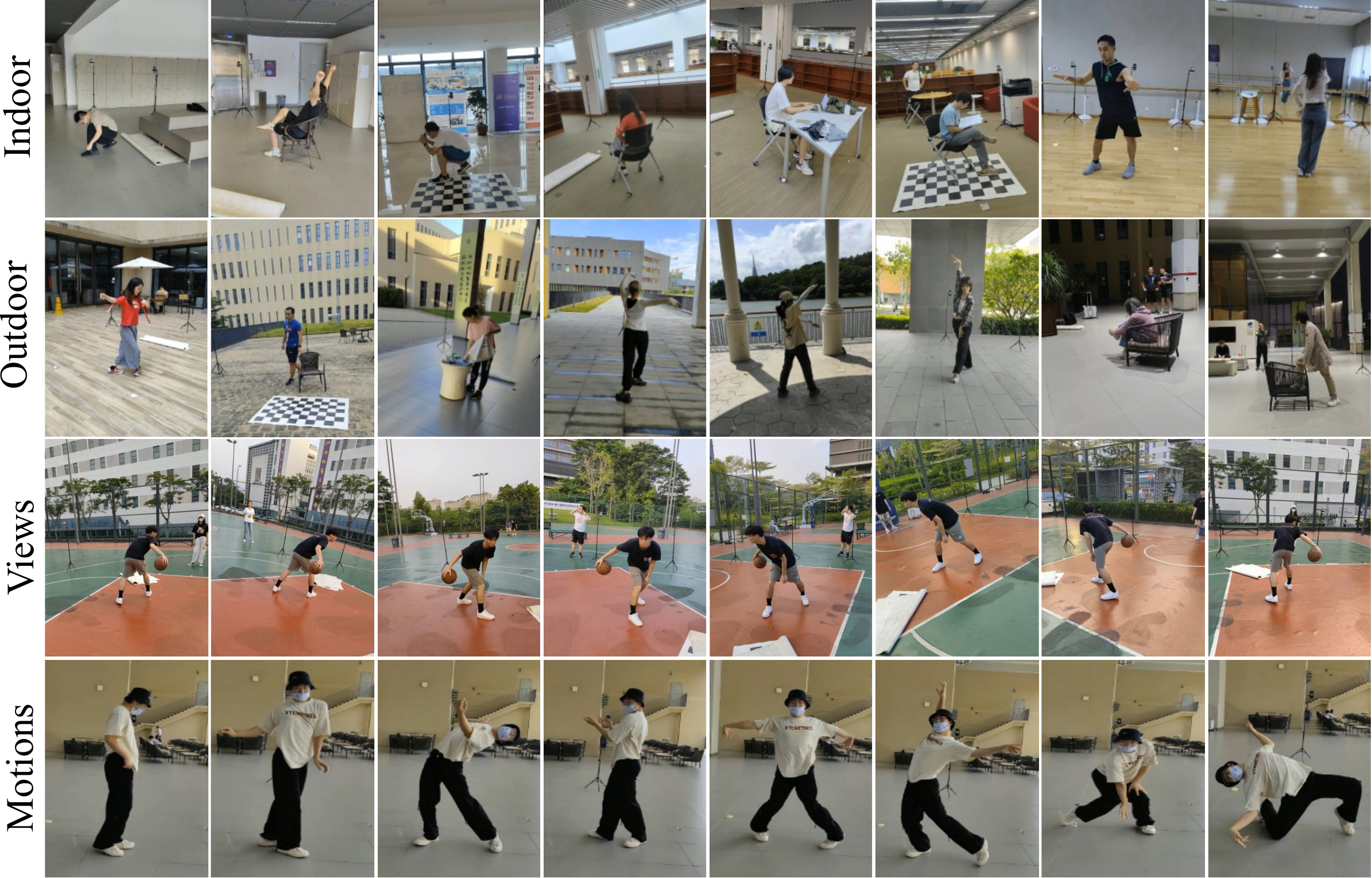}
    \captionsetup{font={small}}
    \caption{The diverse frames in FreeMan. The topmost two rows presents a range of indoor and outdoor scenes, highlighting human-object interactions and the diversity of scene contexts, lighting conditions, and subjects. The third row exhibits frames from different views. The final row illustrates the temporal variation of human poses from a consistent viewpoint, emphasizing the dynamism of motion capture.}
    \label{fig:IntroScenes}
    \vspace{-3mm}
\end{figure*}

\noindent\textbf{Action Set.}
Following the popular action recognition dataset NTU-RGBD120~\cite{liu2019nturgbd}, we compose our action set with several common actions corresponding to scenes in daily life, $e.g.,$ drinking and talking in a cafe, reading in the library. 
Meanwhile, subjects interact with real objects to make data as close to real world as possible.
As shown in the topmost row of Fig.~\ref{fig:IntroScenes}, interaction with objects brings complicated occlusions, making our data more challenging. For outdoor scenarios, we set the data collection field as large as possible to help subjects perform actions with little restriction.

\noindent\textbf{Camera Positions.}
Cameras in previous 3D human pose datasets~\cite{cai2022humman, h36m_pami, Joo_2017_CMU_TPAMI} are fixed, resulting that only a few camera poses being included.
As shown in Fig.~\ref{fig:device_settings}, cameras are attached to tripods and are newly placed from time to time, and translation from the center of the system to camera $d$, which is the physical distance between the camera and the system center, can vary from $2$m to $5.5$m.
Fig.~\ref{fig:dataset_stat} (a) shows the distribution of $d$ and the corresponding number of cameras.
Most cameras are located around $4$ meters far away from the system center. 
Besides, we show the distribution of the human bounding box area in Fig.~\ref{fig:dataset_stat} (b), in a unit of ratio to the whole image area, to demonstrate the variation of human size. 
With variations in camera translation and human actions, the area of human bounding boxes varies from $0.01$ to $0.7$ of the whole image area. 

\noindent\textbf{Subjects.}
There are $40$ subjects participating in the construction of FreeMan and recruitment is completely based on voluntary. 
Among them, 22 actors are trained dancers for dance actions.
All of them are well-informed and signed the agreement to make data public for research purposes only.

\section{Data Acquisition \& Annotation Pipeline}
\label{sec:toolchain}

\textbf{Overview.} To collect a large-scale dataset from real-world environments, we developed a comprehensive toolchain, as shown in Fig.~\ref{fig:toolchain}. Unlike previous toolchains used in controlled or idealized conditions, we carefully accounted for potential challenges in outdoor settings, including calibration and synchronization errors.
To overcome these issues, we proposed an semi-automated pipeline including error detection and manual correction to ensure efficient data collection and annotation.

\subsection{Hardware Setup}
\textbf{Cameras.}
We collect FreeMan via $8$ Mi11 phones~\cite{Mi11_web} indexed from $1$ to $8$ as our data collection devices. 
\textit{\textbf{Note $8$ collection of one action as one session, which corresponds to 8 RGB sequences from 8 views,}} and each phone is attached to a tripod to keep stable during data collection. 
As shown in Fig.~\ref{fig:device_settings}, all devices are positioned in a circle around a human at a height of approximately $1.6$ meters above the ground, and the distance from cameras to the system center varying from $2$ to $5.5$ meters, which is similar to real-life usage scenarios.
Each smartphone captures RGB sequences using its main camera at $1920$$\times$$1080$ resolution and $30/60$ FPS. 
During the data collection process, actors perform actions facing the cameras with odd-numbered indices.
As shown in Fig.~\ref{fig:toolchain} (a), the only requirement beyond devices is a stable network connection to server for data transmission.

\noindent\textbf{Device Synchronization.} 
Previous works~\cite{cai2022humman, h36m_pami, Joo_2017_CMU_TPAMI} have synchronized devices using wired interfaces in a laboratory. 
However, the complexity of the entire system coupled with the difficulty in deploying it in real-world environments, has prompted us to consider alternative methods. 
To address issues related to usability and device constraints, we connect all devices wirelessly to a single server and developed an Android app that utilizes the Network Time Protocol (NTP)~\cite{NetworkTimeProtocol} to calculate the time difference between each device and the server's clock. 
During the capture process, temporal information is stored locally on each device as a timecode, while the server records the synchronized capture interval for all devices. 
The starting frame is determined by matching the timecode to the frame closest to the server's clock time. 
As shown in Fig.~\ref{fig:toolchain}(b), synchronization errors are smaller than a single frame during our testing, corresponding to $33$ms and $16$ms for $30$FPS and $60$FPS, respectively.

\noindent\textbf{Chessboard-based Calibration.}
At the beginning of each session, we first shoot a chessboard with known size tiled at the center of the system, then calculate the intrinsic and extrinsic camera parameters following the standard implementation in OpenCV~\cite{opencv_library, CalibrationZhang}. Please refer to supplementary meterial for details of data flow.


\noindent\textbf{Pixel Alignment}
However, calibration with coarse matching points on chessboard is not accurate enough.
After data collection and synchronization, we extract one frame from all synchronized videos, and then use LightGlue~\cite{lindenberger2023lightglue} to calculate dense matching points across views. 
Then dense matching points are used to further refine the camera extrinsic parameters resulted from chessboard-based calibration.

\subsection{Pose Annotation.}
\begin{figure*}
    \centering
    \includegraphics[width=0.9\linewidth]{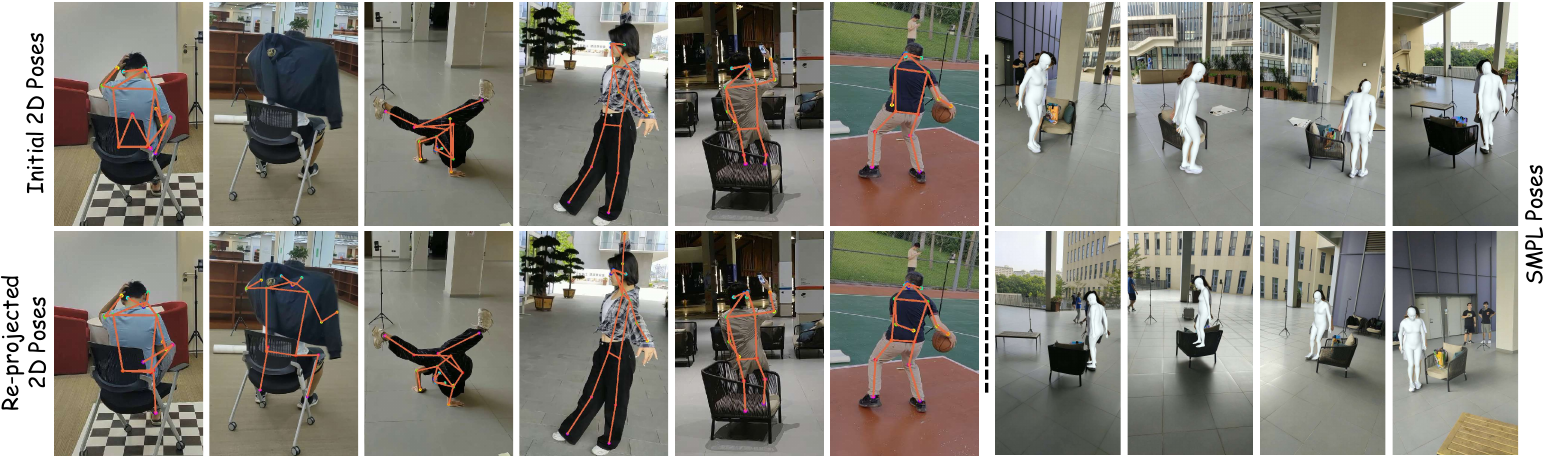}
    \captionsetup{font={small}}
    \caption{
    The examples of human pose annotations are presented as follows. At left, the first row displays 2D keypoints directly generated by HRNet-w48~\cite{sun2019hrnet}, while the second row presents re-projected 2D poses. For heavy occlusions, our pipeline corrects the erroneous keypoints effectively. The right part showcases the SMPL annotation examples for each view in our dataset.
    }
    \label{fig:pose_refine}
\end{figure*}

Once videos are collected, we use a state-of-the-art detector YOLOX~\cite{yolox2021} to detect human bounding box and HRNet-w48~\cite{sun2019hrnet} to detect 2D keypoints of $8$ views ${K}_{2D} \in \mathbb{R}^{8 \times 17 \times 2}$ in COCO~\cite{cocodataset} format. 
To eliminate the effect of potential wrong keypoints output, keypoint predictions with confidence lower than a threshold $\phi$ are removed. 
Then remaining 2D keypoints are used for triangulation to get 3D human pose ${K}_{3D} \in \mathbb{R}^{17 \times 3}$ with pre-computed camera parameters.
Here, we set $\phi$ to be $0.5$.
Furthermore, we optimize ${K}_{3D}$ with smoothness constraints and bone length constraints introduced in HuMMan~\cite{cai2022humman} 
resulting in optimized 3D pose $\Tilde{{K}}_{3D} \in \mathbb{R}^{17 \times 3}$. 
Then we fit a standard SMPL~\cite{SMPL:2015} model to the estimated 3D skeleton by SMPLify~\cite{Bogo_ECCV2016_smplify} to produce a rough mesh annotation.
After that, we project 3D keypoints to 2D image planes of each view using corresponding camera parameters.
With regularization in triangulation and optimization along the temporal axis, the re-projected 2D poses $\Tilde{{K}}_{2D}$ is more accurate than $K_{2D}$, especially for occlusion cases.
Comparison between original $K_{2D}$ and $\Tilde{{K}}_{2D}$ are shown in the left part of Fig.~\ref{fig:pose_refine}.

\begin{figure*}
    \centering
    \includegraphics[width=0.98\linewidth]{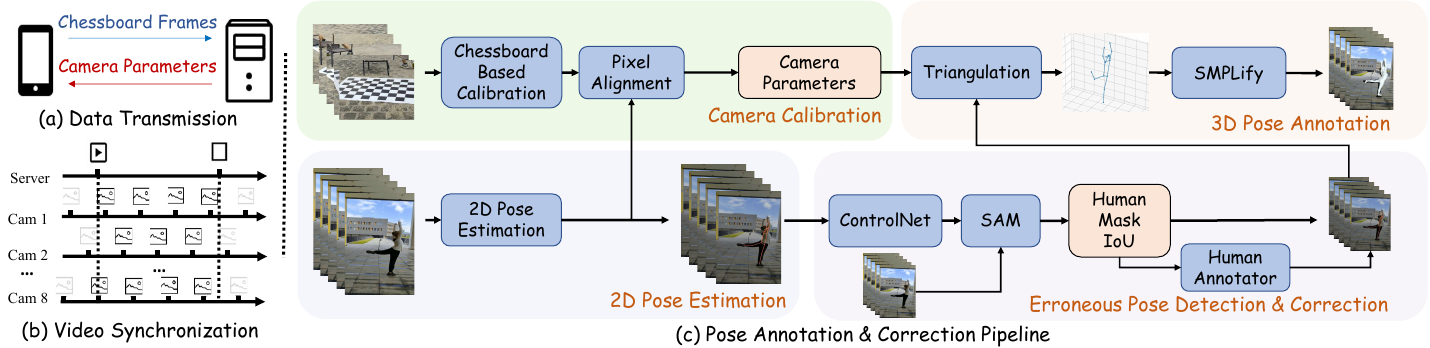}
    \captionsetup{font={small}}
    \caption{
    The illustration of data collection and annotation toolchain: (a) depicts the transmission of signals between cameras and servers for camera calibration, where chessboard frames are sent to the server, and camera parameters are returned. (b) demonstrates the synchronization process among devices. (c) showcases the pipeline for pose annotation.
    }
    \label{fig:toolchain}
\end{figure*}

\begin{figure}
    \centering
    \includegraphics[width=0.9\linewidth]{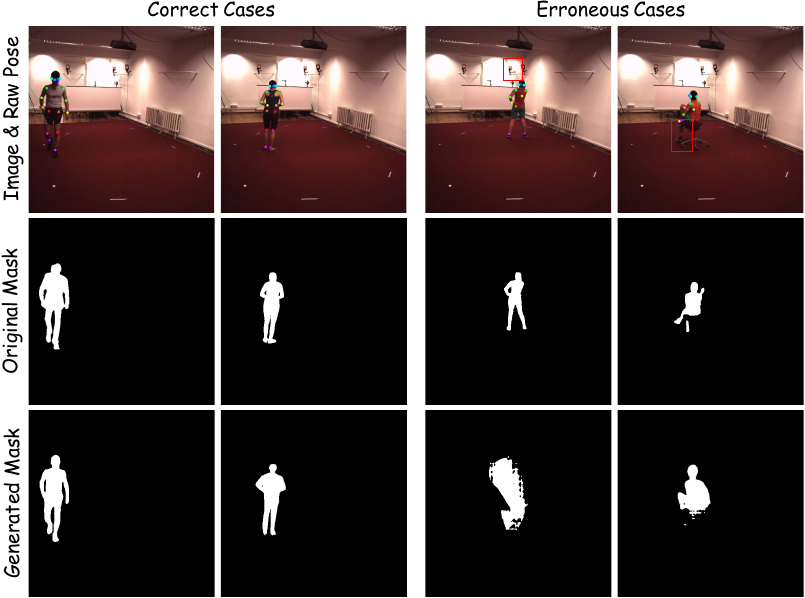}
    \caption{Demonstration of erroneous pose detection in Sec.~\ref{sec:toolchain}. Human3.6M examples shown for quality assessment. The first row shows input frame and 2D keypoints by Pose estimator, and the last two rows show segment mask of original image and generated image by SAM. The left two columns are examples of correct poses, while the right two columns refer to cases with erroneous keypoints as highlighted by the red boxes. Please zoom in for details.}
    \label{fig:error_pose_detect}
    \vspace{-3mm}
\end{figure}

\noindent\textbf{Erroneous Pose Detection \& Correction.} Although 2D pose estimator has been well developed, pose with heavy occlusions can be inaccurate. 
Thus, we propose a pipeline to filter erroneous 2D keypoints among vast millions of frames and then correct them \textbf{by human annotators}.
As shown in Fig.~\ref{fig:toolchain}, estimated 2D poses are feed into a pre-trained image generator to generate human images. 
Then we use SAM~\cite{Kirillov_2023_SAM} to get human mask of original and generated images and intersection-over-union (IoU) between these mask are calculated.
Poses correspond to IoU lower than a threshold $\alpha$ are considered as erroneous ones and then checked by human annotator.
Specifically, we choose Stable Diffusion 1.5 and ControlNet~\cite{Zhang_2023_controlnet} as conditional image generator and DeepDataSpace~\cite{dds_idea_2023} are used as annotation tools.
Fig.~\ref{fig:error_pose_detect} presents examples of correct and erroneous cases. More detailed processes and results are displayed in the supplementary material.

\subsection{Keypoint Quality Assessment} 

To demonstrate the effectiveness of our toolchain, we test it on Human3.6M~\cite{h36m_pami}.
We select $3$ different actions of each subject in the training set, which covers $10\%$ sequences of the whole training set and all kinds of actions. 
Following~\cite{cai2022humman}, keypoint quality is assessed by Euclidean distance between estimated 2D poses and ground truth 2D poses in units of pixels.
The error results in less than 1\% of pixels for images of 1000 × 1000, indicating that our toolchain can generate annotations with an accuracy that is acceptable considering the cognitive errors inherent in human labeling.
\vspace{-3pt}
\section{Benchmarks}

We have constructed four benchmarks utilizing images and annotations derived from our dataset. The data is subdivided based on subjects, allocating $18$ subjects for training, $7$ for validation, and $15$ for testing purposes. This partitioning results in three subsets composed of $5.87M$, $700K$, and $3.69M$ frames, respectively. 
For each benchmark, subject lists of each subset are shared, and only views selected from the session vary for each task.

\noindent\textbf{Monocular 3D Human Pose Estimation (HPE).} 
This task involves taking a monocular RGB image or sequence as input and predicting 3D coordinates in camera coordinate system. 
We randomly select one view from each session for this task.
The performance of algorithms is measured by widely used Mean Per Joint Position Error (MPJPE)~\cite{h36m_pami} and Procrustes analysis MPJPE (PA-MPJPE)~\cite{SMPL:2015}.

\noindent\textbf{2D-to-3D Lifting.}
%
Given that 2D human poses can be predicted using existing 2D keypoint detectors~\cite{cao2017realtime,chen2018cascaded,alphapose,sun2019hrnet}, the primary goal of this task is to effectively elevate these 2D poses into the 3D space within the camera coordinate system. 
The evaluation metrics are the same as HPE.

\noindent\textbf{Multi-View 3D Human Pose Estimation.} 
Estimating the 3D human pose from multiple views presents a natural solution to overcome occulusion in motion capture.
For this task, models are provided with images or videos from multiple views, along with corresponding camera parameters. The objective is to predict the 3D coordinates of human joints in the same world coordinate system as the cameras.
following implementation of~\cite{voxelpose}, metrics of the task is MPJPE and average precision (AP) with specific thresholds.
%

\noindent\textbf{Human Neural Rendering.} 
The free-viewpoint rendering of humans is a significant issue in human modeling. With the rise in popularity of neural radiance fields (NeRF)~\cite{mildenhall2021nerf} for the novel view rendering task, several methods, including HumanNeRF~\cite{weng2022humannerf}, have emerged. These methods utilize monocular human motion videos as input to synthesize novel views of dynamic humans through NeRF.
The widely used metrics of prediction are PSNR, SSIM~\cite{zhang2021nerfactor} and LPIPS~\cite{zhang2018unreasonable}.

\vspace{-3mm}
\section{Experiments}
\vspace{-2mm}
In this section, we experiment with the four benchmarks. In human 3D pose estimation tasks, we conduct several transfer tests with other standard datasets to evaluate the effectiveness and transferability of our proposed FreeMan dataset. Existing similar datasets, Human3.6M~\cite{h36m_pami} \& HuMMan~\cite{cai2022humman}, are used for comparison. 
Since \textit{HuMMan only releases 1\% of data for pose estimation}\footnote{\url{https://opendatalab.com/OpenXDLab/HuMMan}}, we only involve it in monocular 3D human pose estimation and 2D-to-3D lifting.
As for the neural human rendering, we train the model from one of the $8$ views and test on the other the views in selected sessions.

\subsection{Monocular 3D Human Pose Estimation}

\begin{table}[t]
\centering
    \begin{adjustbox}{max width=\linewidth}
    \begin{tabular}{ccc|cc|cc}
    \toprule[1.5pt]
         \multicolumn{3}{c}{Method} & \multicolumn{2}{c}{\textbf{HMR}} & \multicolumn{2}{c}{\textbf{PARE}} \\
         \hline
         Train & Supervision & Test & MPJPE & PA & MPJPE & PA\\ 

         \hline
         Human3.6M & 2D+3D KPTs & 3DPW & 279.92 & 133.13 & 118.54 & 81.22  \\
         HuMMan & 2D+3D KPTs & 3DPW & 407.57 & 192.75 & 110.99 & 63.11 \\
         HuMMan & 2D KPTs+SMPL &3DPW & 475.73 & 184.15 & 114.20 & 66.19 \\
         HuMMan & 2D+3D KPTs+SMPL & 3DPW & 437.52 & 203.17 & 114.33 & 72.12  \\ \hline 
         FreeMan & 2D+3D KPTs & 3DPW & 157.46 & 87.93\color{Green}$\uparrow_{33.95\%}$ & 118.31 & 68.72\color{Green}$\uparrow_{15.39\%}$ \\
         FreeMan & 2D KPTs+SMPL & 3DPW & 151.85 & 88.85\color{blue}$\uparrow_{51.75\%}$ & 94.27 & 60.39\color{blue}$\uparrow_{8.76\%}$ \\
         FreeMan & 2D+3D KPTs+SMPL & 3DPW & 159.31 & 91.33\color{blue}$\uparrow_{55.04\%}$ & 98.33 & 64.51\color{blue}$\uparrow_{10.55\%}$ \\
    \bottomrule[1.5pt]
    \end{tabular}
    \end{adjustbox}
    \captionsetup{font={small}}
    \tabcaption{Monocular 3D HPE performance of HMR~\cite{hmrKanazawa17} and PARE~\cite{DBLP:journals/corr/abs-2104-08527} trained on different dataset for monocular Human Pose Estimation. PA stands for PA-MPJPE and both metrics are in unit of millimeters. The lower metrics is, the better performance model obtains. All released part of HuMMan is used for training. {\color{blue}{$\uparrow$}} refers to the improvement relative to HuMMan and {\color{Green}$\uparrow$} refers to the improvent relative to Human3.6M.
    }
    \label{tab:monoHPE}
    \vspace{-3mm}
\end{table}

\noindent\textbf{Implementation details.} 
For the Human3.6M~\cite{h36m_pami} and HuMMan~\cite{cai2022humman} datasets, all views in their training set are utilized. 
To balance number of frames, we randomly sample a single view from sessions in the training split for FreeMan, resulting that the frame numbers of all three datasets being $312K$, $253K$, and $233K$, respectively.
Videos of all three datasets are downsampled to $10$FPS, following the implementation of MMPose~\cite{mmpose2020}.
Following~\cite{Benchmark}, we select HMR~\cite{hmrKanazawa17} and PARE~\cite{DBLP:journals/corr/abs-2104-08527} as models to evaluate and implement experiments with configurations open-sourced by~\cite{Benchmark}.
Please refer to supplementary material for more.

\noindent\textbf{Results.} 
We perform testing on the test set of 3DPW~\cite{vonMarcard2018}. The performance of the models trained on different datasets, with varying types of supervision, are reported in Tab.~\ref{tab:monoHPE}.
Notably, the HMR models trained on FreeMan exhibit significantly better performance on the 3DPW test set compared to those trained on Human3.6M and HuMMan with PA-MPJPE $133.13$mm and $192.75$mm respectively, which indicates that FreeMan demonstrates superior generalizability compared to the others.
The same results are obtained with PARE, further confirming that FreeMan outperforms even in more advanced algorithms. 
This can be attributed to the diversity of scene contexts and human actions present in our dataset, which provides better transferability in real-world scenarios.

\subsection{2D-to-3D Pose Lifting}

\begin{table}
    \centering
    \begin{adjustbox}{max width=\linewidth}
        \begin{tabular}{c | c c c c}
        \toprule[1.5pt]
        \multirow{1}{*}{Algorithm} &\multirow{1}{*}{Train} &
        \multirow{1}{*}{Test} &
        \multirow{1}{*}{MPJPE (mm)} &\multirow{1}{*}{PA (mm)} \\ \hline
        \multirow{5}{*}{SimpleBaseline} & FreeMan & FreeMan & 90.53 & 54.17 \\ 
        & FreeMan$^{\dag}$ & FreeMan &79.22 & 49.11 \\ 
        &Human3.6M &AIST++ &212.57 &138.98 \\ 
        &HuMMan &AIST++ &255.5 &116.86 \\ 
         & \cellcolor{gray!10}FreeMan& \cellcolor{gray!10}AIST++ & \cellcolor{gray!10}156.96 & \cellcolor{gray!10}105.85\color{Green}$\uparrow_{10.30\%}$\\ 
        &\cellcolor{gray!20}FreeMan$^{\dag}$ &\cellcolor{gray!20}AIST++ &\cellcolor{gray!20}126.23 &\cellcolor{gray!20}88.07\color{Green}$\uparrow_{24.64\%}$ \\  \hline
        
        \multirow{5}{*}{MHFormer} & FreeMan & FreeMan & 93.00 & 63.50 \\ 
        & FreeMan$^{\dag}$ & FreeMan &77.06 & 53.38 \\ 
        &Human3.6M &AIST++ &171.19 &133.37 \\ 
        &HuMMan &AIST++ &188.73 &101.52 \\ 
        & \cellcolor{gray!10}FreeMan & \cellcolor{gray!10}AIST++ & \cellcolor{gray!10}132.99 & \cellcolor{gray!10}88.79\color{Green}$\uparrow_{12.54\%}$\\ 
        &\cellcolor{gray!20}FreeMan$^{\dag}$ &\cellcolor{gray!20}AIST++ &\cellcolor{gray!20}124.34 &\cellcolor{gray!20}79.22\color{Green}$\uparrow_{21.97\%}$ \\
        \bottomrule[1.5pt]
        \end{tabular}
    \end{adjustbox}
    \caption{Performance of methods with different training and testing datasets in 2D-to-3D Pose Lifting. PA stands for PA-MPJPE. 
    $^{\dag}$ refer to experiments with the whole training set of FreeMan. Smaller MPJPE and PA-MPJPE indicate better performance.
    Highlighted rows show training on our dataset achieves the best performance in the transfer test. {\color{Green}$\uparrow$} refers to the improvement relative to HuMMan.} 
    \label{tab:poselift}
    \vspace{-3mm}
\end{table}

\begin{table*}[th]
    \centering
    \begin{adjustbox}{max width=0.9\textwidth}
    \begin{tabular}{cccccccc}
    \toprule[1.5pt]
         Train & Test & AP@25mm (\%) $\uparrow$ & AP@50mm (\%) $\uparrow$ & AP@75mm (\%) $\uparrow$& AP@100mm (\%) $\uparrow$ & Recall@500mm (\%) $\uparrow$ & MPJPE@500mm (mm) $\downarrow$ \\ 
         \hline
         Human3.6M & Human3.6M & 32.32 & 97.47 & 98.61 & 98.99 & 100.00 & 25.29 \\
         Human3.6M & FreeMan & 0.00 & 0.00 & 0.00 & 0.00 & 0.06 & 89.85 \\
         Human3.6M & FreeMan (w/ GT Root) & 0.00 & 1.27 & 11.44 & 21.40 & 96.20 & 103.02 \\
         \hline
         FreeMan & FreeMan  & 43.38 & 88.77 & 97.73 & 99.12 & 99.97 & 26.07 \\
         \rowcolor{gray!10}FreeMan & Human3.6M  & 0.00 & 5.77 & 82.85 & 92.62 & 96.68 & 61.29 \\
         \rowcolor{gray!20}FreeMan & Human3.6M (w/ GT Root) & 0.00 & 6.60 & 87.91 & 95.38 & 100.00 & 58.30\\
    \bottomrule[1.5pt]
    \end{tabular}
    \end{adjustbox}

    \caption{Multi-View 3D Pose Estimation results of VoxelPose~\cite{voxelpose}. 
    Ground truth root position (GT Root) is not used if not specified. 
    Recall@$500mm$ shows the percentage that falls within the threshold, and the MPJPE@$500mm$ indicates the average MPJPE values within the threshold.
    Rows highlighted shows the best setting in cross-domain test.}
    \label{tab:multiview_500mm}
    \vspace{-3mm}
\end{table*}

\begin{table}[th]
\centering
\scalebox{0.75}{
\begin{tabular}{c|ccc}
    \toprule[1.5pt]
         Scene & PSNR$\uparrow$ & SSIM$\uparrow$ & LPIPS$^{*}\downarrow$ \\ 
         \hline
         Square & 25.98 & 0.9501 & 58.38  \\
         Corridor & 24.57 & \underline{0.9340} & \textbf{81.39} \\
         Sports Port & 26.33 & 0.9662 & 30.09  \\
         Park & \underline{23.86} & 0.9439 & 73.61  \\
         Courtyard & 28.56 & 0.9630 & 53.99  \\
         Dance Room & \textbf{30.11} & 0.9658 & 43.34  \\
         Library & 29.41 & \textbf{0.9665} & \underline{31.53}  \\
         Platform & 26.79 & 0.9439 & 70.01  \\
         Lobby & 25.41 & 0.9387 & 78.80  \\
         Cafe & 27.32 & 0.9644 & 37.88  \\
    \bottomrule[1.5pt]
\end{tabular}}
\caption{Neural rendering results by using HumanNeRF~\cite{weng2022humannerf} on $10$ scenes of FreeMan. Note that \texttt{LPIPS$^{*}$} = \texttt{LPIPS} $\times 10^3$. The highest values are bolded and underlined ones refer to the lowest.}
\label{tab:humannerf}
\end{table}

\noindent\textbf{Implementation Details.}  
For this task, we employ CNN-based methods, SimpleBaseline~\cite{DBLP:journals/corr/MartinezHRL17} and VideoPose3D~\cite{DBLP:journals/corr/abs-1811-11742}, and Transformer-based methods, PoseFormer~\cite{DBLP:journals/corr/abs-2103-10455} and MHFormer~\cite{DBLP:journals/corr/abs-2111-12707}, and all methods follow corresponding official implementations.
To verify the effect of the dataset scale, we also train our model on the whole training set.
The results of SimpleBaseline and MHFormer are presented in Tab.~\ref{tab:poselift}, and more details can be found in supplementary material.

\noindent\textbf{Results.}
As shown in Tab.~\ref{tab:poselift}, results of the in-domain test on FreeMan are provided as a baseline for future work.
For in-domain testing, MPJPE of SimpleBaseline on FreeMan ($79.22$mm) is larger than that on HuMMan~\cite{cai2022humman} ($78.5$mm\footnote{As full data not accessible, we use result from HuMMan~\cite{cai2022humman} directly.}) and Human3.6M~\cite{h36m_pami} ($53.4$mm\footnote{Results of our implementation.}), demonstrating that FreeMan is a more challenging benchmark.
Besides, all the methods trained on FreeMan tend to generalize better than that on HuMMan and Human3.6M when testing on AIST++ under the same setting as MPJPE and PA-MPJPE are much smaller in cross-domain test. 
Although the scale of FreeMan training set is of a similar magnitude as HuMMan's, which is much smaller than Human3.6M's, models trained on FreeMan outperform models trained on the other two by a large margin.
Furthermore, when the training set is expanded to all frames in training split, FreeMan can further boost models to achieve better performance, proving that our large-scale data helps to improve model performance.

\subsection{Multi-View 3D Human Pose Estimation} 

\noindent\textbf{Implementation Details.}
We conduct in-domain and cross-domain tests between Human3.6M and FreeMan to evaluate the effectiveness and generalization ability. 
We conduct the experiments with VoxelPose~\cite{tu2020voxelpose}, which locates the human root first and then regresses 3D joint location accordingly.
COCO-format poses in FreeMan are interpolated to match that in Human3.6M.
We trained VoxelPose~\cite{tu2020voxelpose} following official implementation.
For Human3.6M, bounding box annotations are from~\cite{qiu2019cross} and its validation set is used for the test.
For FreeMan, we only use 4 odd-indexed views from the training set.

\noindent\textbf{Results.}
Results of all experiments are reported in Tab.~\ref{tab:multiview_500mm}. 
For in-domain testing, the model trained on FreeMan achieves MPJPE@$500$mm of $26.61$mm on test set consisting of \textit{odd-indexed} views.
For cross-domain testing, the model trained on FreeMan achieves Recall@$500$mm of $96.68$\% and MPJPE@$500$mm is $61.29$mm on Human3.6M validation set.
However, the model trained on the Human3.6M dataset fails to locate human on FreeMan test set, resulting zero AP with threshold smaller than 100mm.
To get rid of the effects of root location, we input the ground truth root locations to model directly. 
With this setting, the model trained on Human3.6M obtains MPJPE@$500$mm of $103.02$mm on FreeMan test set, while the model trained on FreeMan can obtain MPJPE@$500$mm of $58.30$mm on Human3.6M validation set. 
Results show that the model trained on FreeMan has a much better generalization ability, while that on Human3.6M struggles in transfer testing.

\subsection{Neural Rendering of Human Subjects.}
\noindent\textbf{Implementation Details.}
We employ $10$ scenes captured by FreeMan to train HumanNeRF~\cite{weng2022humannerf}.
To obtain human body segmentation annotations, we utilize the SAM~\cite{Kirillov_2023_SAM} algorithm with our bounding boxes as prompts. 
Throughout the training step, we randomly select one view for each session and render the rest $7$ view as novel views for testing. 
We then calculate metrics including PSNR, SSIM, and LPIPS, to evaluate the performance of the model. 
Please refer to supplementary material for results of data at $60$FPS.

\noindent\textbf{Results.}
The reconstruction results in $10$ scenes are shown in Tab.~\ref{tab:humannerf}. The best reconstruction achieves a high PSNR of $30.11$dB which indicates FreeMan contains contents that the models can learn and fit very well. While the performance varies, the lowest PSNR of $23.86$ shows FreeMan also contains contents that are outside of model's learning scope and challenging. Additionally, the results in $10$ scenes including both easy contents that the model can handle well and challenging contents demonstrating the diversity of FreeMan.

\vspace{-2mm}
\section{Conclusion}
We present FreeMan, a novel large-scale multi-view 3D pose estimation dataset 3D human pose annotations. 
We elaborately develop a simple yet effective semi-annotation pipeline to automatically annotate frame-level 3D landmarks at a much lower cost, and build a comprehensive benchmark for 3D human pose estimation.
%

Extensive experimental results demonstrate the difficulty of test in varied conditions and the strengths of the proposed FreeMan. 
As a large-scale human motion dataset, our FreeMan addresses the existing gap between the current datasets and real-scene applications, and we hope that it will catalyze the development of algorithms designed to model and sense human behavior in real-world scenes.

\textbf{Limitations.} 
Prompts to generation model require careful tuning for high quality and accuracy of error pose detection can be limited by human image generation models. 


\vspace{-2mm}
\section*{Acknowledgement}
We sincerely thank all volunteers and MaxDancingClub from CUHK(SZ) for participation, and Mr. Ruipeng Cao for software development.
The work is partially supported by the Young Scientists Fund of the National Natural Science Foundation of China under grant No. 62106154, by the Natural Science Foundation of Guangdong Province, China (General Program) under grant No.2022A1515011524, and by Shenzhen Science and Technology Program JCYJ20220818103001002 and ZDSYS20211021111415025 and by the Guangdong Provincial Key Laboratory of Big Data Computing, The Chinese University of Hong Kong (Shenzhen).

{
    \small
    \bibliographystyle{ieeenat_fullname}
    \bibliography{main}
}

\clearpage
\setcounter{page}{1}
\setcounter{section}{0}
\maketitlesupplementary



\begin{figure*}[t]
    \centering
    \includegraphics[width=\textwidth]{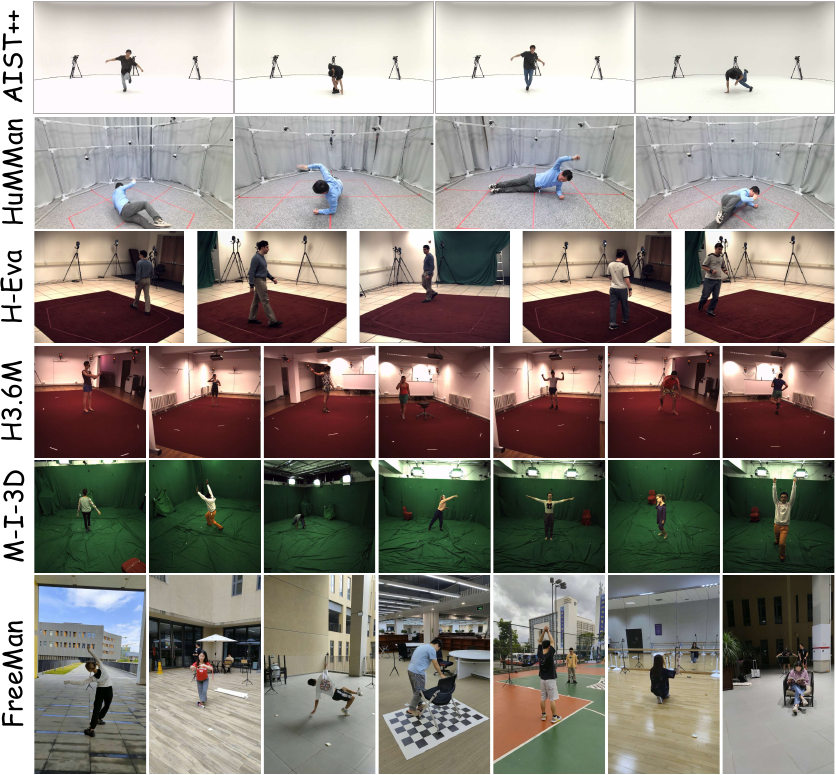}
    \caption{Comparison between FreeMan and existing multi-view 3D human datasets, and all images maintain their original aspect ratio. H-Eva, H3.6M and M-I-3D represents Human-Eva~\cite{humaneva/ijcv/SigalBB10}, Human3.6M~\cite{h36m_pami} and MPI-INF-3DHP~\cite{mono-3dhp2017}, respectively. Datasets except for FreeMan are all collected in fixed laboratory environment. Although MPI-INF-3DHP composites different textures to the background, its human-object interaction and action sets are much simpler.}
    \label{fig:dataset_comparison}
\end{figure*}

\begin{figure*}[ht]
    \centering
    \includegraphics[width=\textwidth]{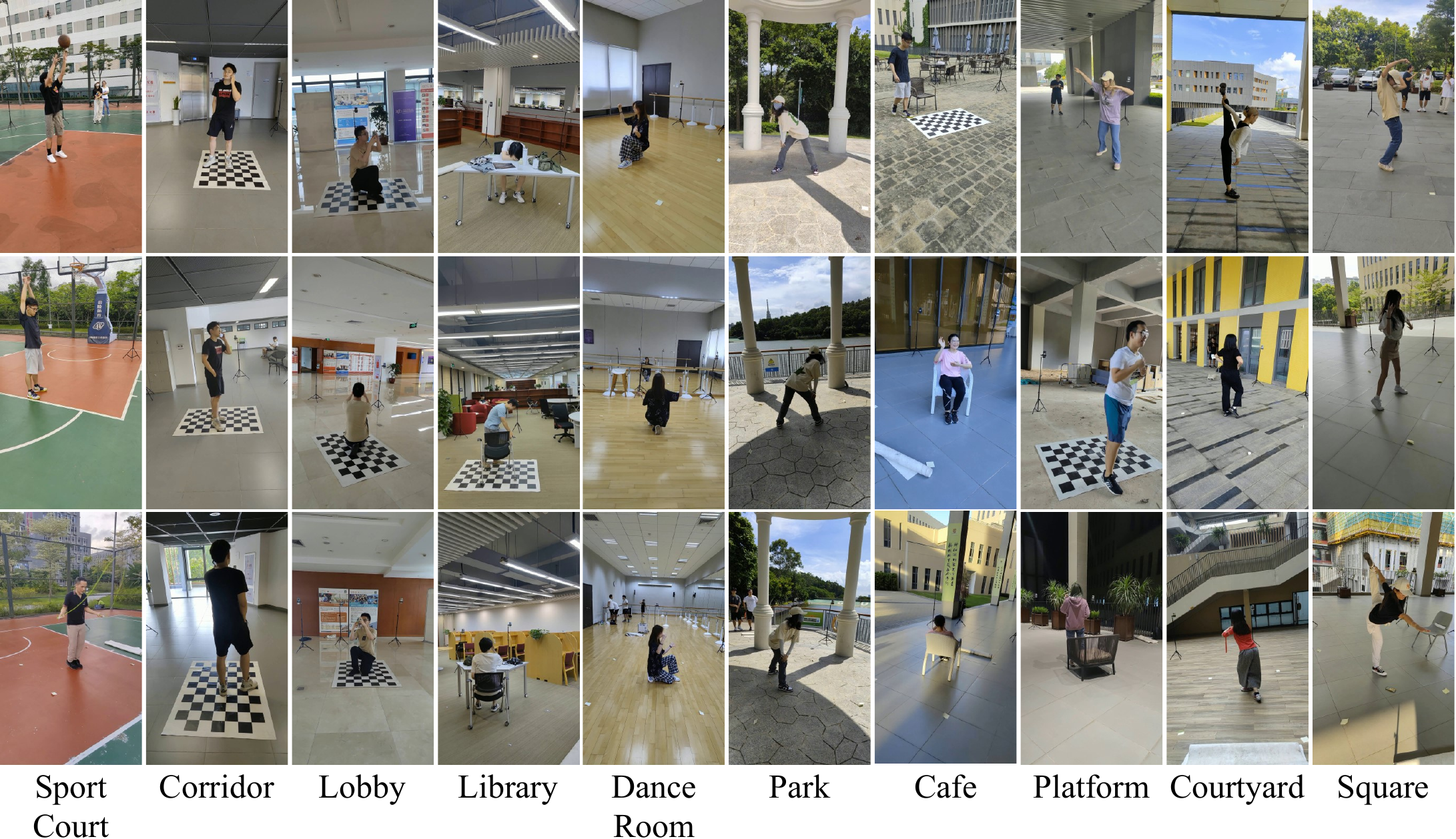}
    \caption{Example images of 10 kinds of scenarios. For scenes shown in leftmost six columns, various of background from different views are presented. For outdoor scenes such as cafe, platform, courtyard and square, \textbf{various locations are included}.}
    \label{fig:SceneExample}
\end{figure*}

\begin{figure}[ht]
    \centering
    \includegraphics[width=\linewidth]{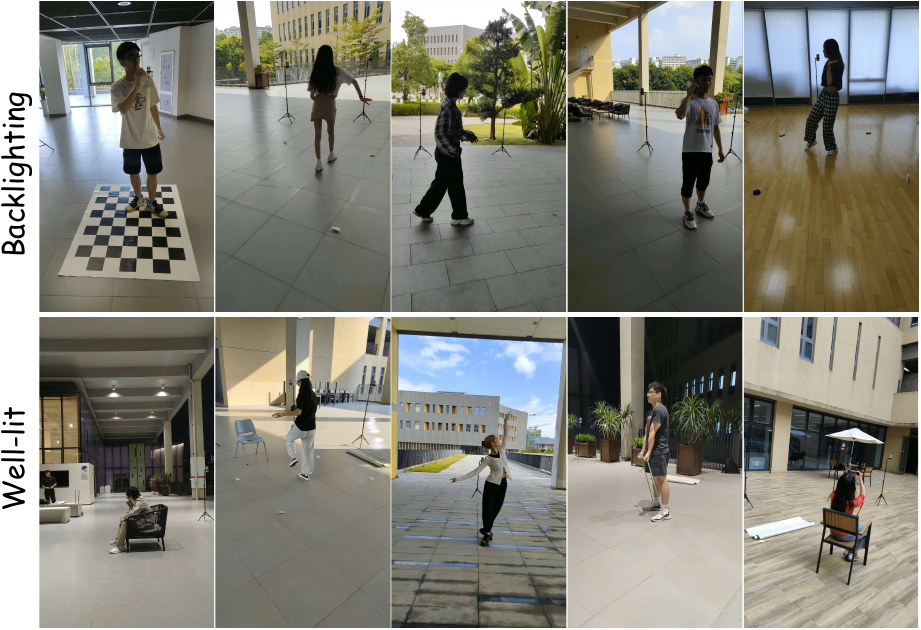}
    \caption{Example images for lighting conditions. The first row presents cases of backlit, resulting in reduced visibility and a relatively darker appearance of the person. The second row shows cases of well-lit, which is normal cases in real world.}
    \label{fig:lighting}
\end{figure}

\begin{figure*}[th]
    \centering
    \includegraphics[width=\textwidth]{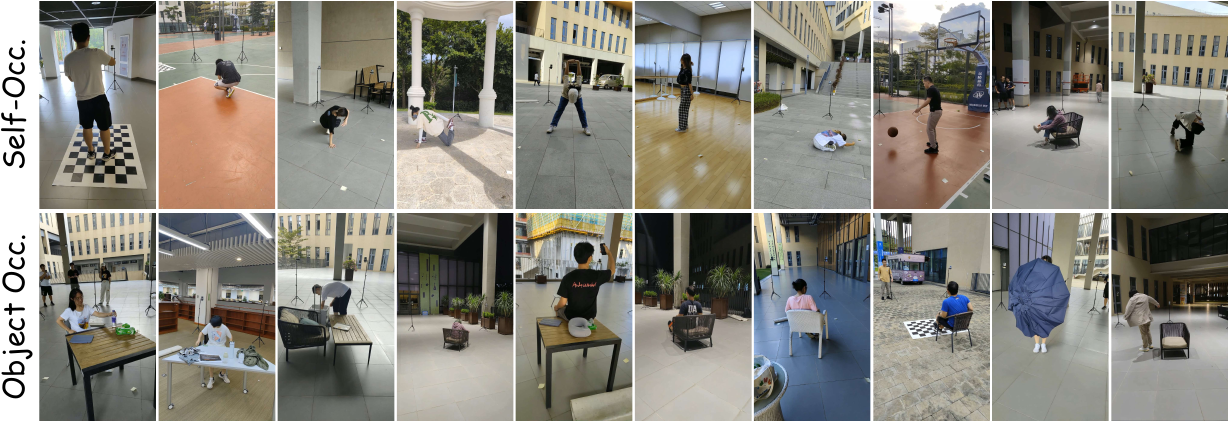}
    \caption{Examples of occlusion in FreeMan. Top row shows self-occlusion and the bottom row presents occlusions in human-object interactions.}
    \label{fig:occlusion}
\end{figure*}

\section{Overview}
FreeMan consists of data from 10 types of scenarios and 27 locations, and lighting conditions are various among locations. 
Meanwhile, we adopt different action sets for scenarios to make FreeMan more diverse.
In this section, we first compare FreeMan with existing relevant works to visually demonstrate its diversity. 
Then, we showcase more examples from various perspectives, including scenes, lighting conditions, and occlusions, to provide a more comprehensive representation of FreeMan.

\subsection{Comparison with Existing Datasets}
By showing example frames in Fig.~\ref{fig:dataset_comparison}, we demonstrate the breadth and realism of our 3D human body dataset, which surpasses previous works by encompassing diverse and dynamic real-world conditions. 
We believe that our dataset will serve as a valuable resource for advancing research in computer vision, human pose estimation, and action recognition.

\subsection{Scenes}
In this section, we provide an overview of the $10$ scene categories included in FreeMan. 
Fig.~\ref{fig:SceneExample} presents three images for each scenario categories.
It can be seen that background are much different across scenarios.
For each category, we present three representative images that capture the essence of the scene and highlight the diversity of actions performed.

\subsection{Lighting}
This section focuses on the different lighting conditions captured in our dataset. 
We specifically address challenging lighting scenarios such as backlighting and overexposure, which are common in real-world environments. 
We include a collection of images in Fig.~\ref{fig:lighting} that demonstrate how FreeMan represents these variations and challenges in lighting.
Besides well-lit cases, there are also challenging cases of backlighting, which is also common in real world.

\subsection{Occlusion}
The third section highlights human occlusion phenomena, encompassing both object interactions and self-occlusion during complex actions as shown in Fig.~\ref{fig:occlusion}.
We provide visual examples showcasing instances where the human body is partially or fully occluded by objects, as well as instances where self-occlusion occurs during intricate movements.

\section{Limitations \& Future Work}

FreeMan is the first attempt to address the challenges of 3D human pose estimation in real-world environments with diverse variations. 
However, it is important to acknowledge that the real world encompasses a multitude of variables, and there are additional conditions that can be further explored. Despite the limited size of FreeMan, it serves as a catalyst for advancing algorithmic research in this domain and provides a means to evaluate the performance of existing methods under varying conditions.

Currently, our pose annotations are in the form of 17 keypoints following the COCO format. 
However, given the increasing demand for fine-grained human body modeling, the estimation of whole-body key points has become more crucial. 
In future work, we can consider expanding our pose annotations to cover the entire body, enabling more comprehensive analysis and modeling.

In addition to pose estimation, we have also extended FreeMan dataset to improve the performance of human body rendering algorithms in real-world scenarios. 
Leveraging the data collected under diverse and dynamic environmental conditions, FreeMan offers a more realistic input for studying the robustness of human body rendering algorithms. 
Furthermore, in addition to NeRF-based algorithms~\cite{weng2022humannerf}, future research can explore the application of recent dynamic 3D Gaussian splatting algorithms~\cite{luiten2023dynamic} with FreeMan dataset.


\section{Toolchain}
In this section, we introduce more details of our toolchain. 
this section focus on pixel alignment for camera calibration and erroneous pose detection.

\subsection{Camera Calibration}
To calibrate camera accurately, FreeMan adopts two-stage camera calibration.
Before collecting each session, we collect frames of chessboard and conduct standard calibration process in OpenCV~\cite{opencv_library, CalibrationZhang}.
At this stage, we first calibrated the cameras by capturing at least thirty frames of a calibration board with each camera to determine their intrinsic parameters. Next, the cameras were fixed in a stationary position on tripods, and all cameras simultaneously captured images of a calibration board at the central position to calculate the extrinsic parameters. 
However, due to variations in lighting conditions and distances from certain perspectives, errors in estimating the extrinsic parameters are inevitable in this step.

To address the errors in extrinsic parameter calibration, we introduce pixel alignment and calculate dense image matching points using synchronized video content. 
By leveraging well-synchronized video data, we can establish correspondence between pixels in different camera views, allowing for more accurate estimation of the extrinsic parameters. 

We use LightGlue~\cite{lindenberger2023lightglue} to calculate image correspondances between adjacent views.
For all 8 viewpoints, we take each viewpoint as the center and select three consecutive viewpoints as a group.
For each group, the pixel matcher calculates matching pairs of pixels between two adjacent views and then filters out the pairs of pixels that are common to the three views. For each set of videos, we select frames containing at least 50 sets of pixel pairs, and then use these pixel pairs to calculate camera extrinsic parameters.
Once this process is completed for all viewpoints, we perform coordinate system matching to align all the viewpoints within the same world coordinate system.

\begin{figure*}
    \centering
    \includegraphics[width=\textwidth]{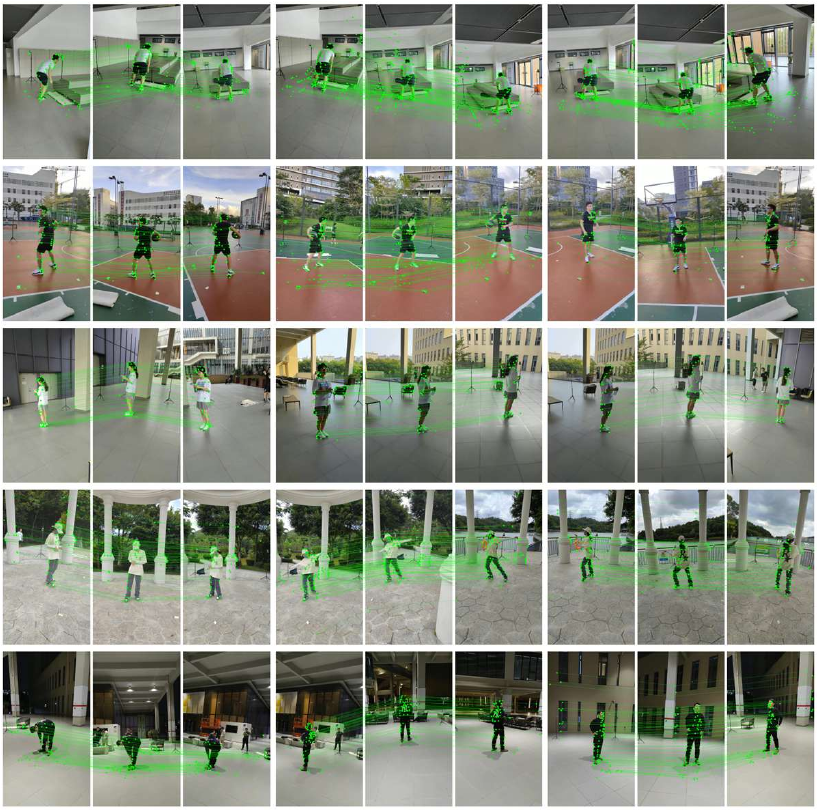}
    \caption{Example of calculated image correspondances of 5 scenes. Three adjacent viewpoints form one group, and three groups for each scenario .}
    \label{fig:pixel_align}
\end{figure*}

\subsection{Erroneous Detection \& Correction}
As presented in Sec. 4.2 in paper, we propose an error detection pipeline and then correct detected frames manually to deal with potential error from 2D pose estimator efficiently.

\begin{figure*}
    \centering
    \includegraphics[width=0.9\textwidth]{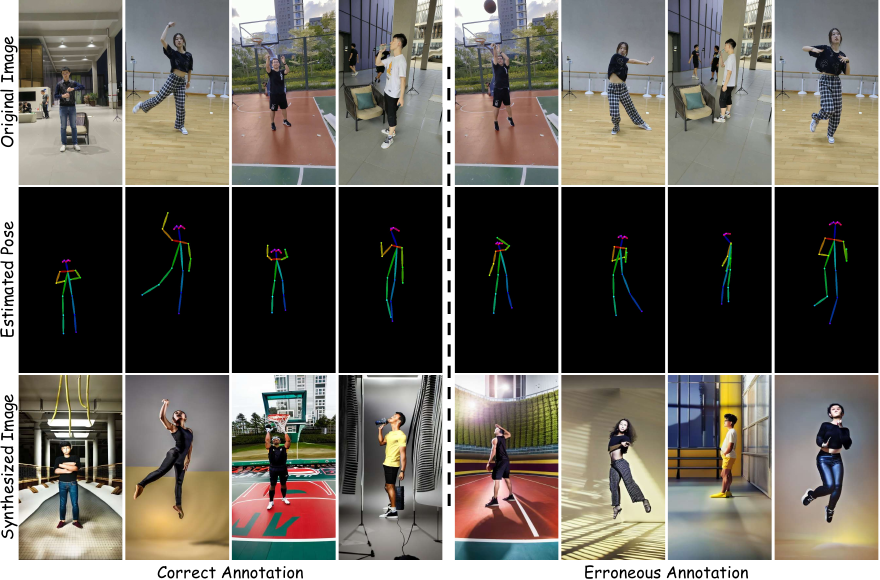}
    \caption{Results of erroneous pose detection. Original images, estimated pose by 2D pose estimator and synthesized images are presented in three rows, respectively. The left part shows cases of correct pose annotation, while the right part presents cases of erroneous poses. 
    In the synthesized image, erroneous pose annotation usually results a completely different body part from the corresponding part in the original image, which means lower IoU (Intersection over Union) for the human body mask compared to the correct annotation.}
    \label{fig:Appendix_error_pose_detect}
\end{figure*}

Given an estimated human pose, we feed it to a pre-trained conditional image generator, ControlNet~\cite{Zhang_2023_controlnet} with Stable Diffusion v1.5~\cite{Rombach_2022_CVPR}. 
Original frame is also applied as a condition for synthesis by DDIM inversion~\cite{mokady2023null}. Besides, scenario category, actions and brief description of actor in each session are input as text prompts.
Then we use SAM~\cite{Kirillov_2023_SAM} to process synthesized and original images and obtain binary human masks, using keypoints as prompts.
If intersection-over-union (IoU) between the two masks is lower than pre-defined threshold, corresponding pose is classified as erroneous one and then will be correct by human annotator. 
Fig.~\ref{fig:Appendix_error_pose_detect} present results of correct and erroneous poses.
We use implementation of Diffuser~\cite{von-platen-etal-2022-diffusers} for image generator and official implementation of SAM. 
Notably, as ControlNet expects poses in OpenPose format~\cite{openpose}, we transfer COCO-format annotations to OpenPose and plot the skeleton image in corresponding color pattern. Neck in OpenPose skeleton is defined to be the mid-point between shoulder keypoints.
Moreover, we use DeepDataSpace~\cite{dds_idea_2023} as manual annotation tool which supports annotation by dragging keypoints.

\section{Experiments}

In this section, we present more details about experiments of benchmarks we set and provide further results of extensive experiments. 

\subsection{Monocular 3D Human Pose Estimation}

For training of HMR~\cite{hmrKanazawa17}, we use Adam optimizer with fixed learning rate of ${2.5 \times 10^{-4}}$. 
Training processes are conducted on a single NVIDIA RTX-3090-24GB GPU with batch size of $128$. 
Additionally, for training of PARE~\cite{DBLP:journals/corr/abs-2104-08527}, we use Adam optimizer with fixed learning rate of ${5.0 \times 10^{-5}}$ at the backbone and head of the network. Training processes are conducted on a single NVIDIA RTX-3090-24GB GPU with batchsize of $128$. Only 2D and 3D keypoints in FreeMan are converted to the format of \textit{HumanData} provided in mmhuman3D~\cite{mmhuman3d}. 
And for finetuning of PARE, we use Adam optimizer with fixed learning rate of ${1.0 \times 10^{-5}}$. The training process are conducted on a single NVIDIA RTX-3090-24GB GPU with batchsize of 128.

We conduct cross-dataset test on FreeMan, Human3.6M and HuMMan, the results of HMR on all test sets are shown in the Tab.~\ref{tab:cross_results}. 
When testing on Human3.6M, in-domain test obtains the lowest error while models trained on FreeMan achieves the second place (192.19mm) and surpass model trained on Human3.6M (465.1mm) greatly.
For HuMMan test set, model trained on FreeMan still achieves better performance than than on Human3.6M. Since some recent work~\cite{DBLP:journals/corr/abs-2104-08527} improves models' performance through dataset mixture, we further finetune the pre-trained PARE model, and results are shown in Tab. \ref{tab:rebuttal_pare_exp}. It can be seen that the 3D HPE performance improvement by fintuning on FreeMan is still higher than HuMMan.

We believe that the improvements mentioned above are due to the diversity of the FreeMan dataset mentioned in dataset overview, which makes the model more robust and generalizable. 
At the same time, the noticeably higher MPJPE on the FreeMan test set compared to other test sets indicates that the FreeMan is also a challenging benchmark.

\begin{table}[!t]
    \centering
    
    \begin{adjustbox}{max width=\linewidth}
    \begin{tabular}{c|ccc}
    \hline
    \diagbox{Train}{MPJPE/PA-MPJPE(mm)}{Test} & H36M & HuMMan & FreeMan\\ \hline
    H36M & 98.62/59.17 & 392.89/175.94 & 350.97/178.85\\ \hline
    HuMMan & 465.1/224.53 & - & 413.26/218.28\\ \hline
    FreeMan & 192.19/112.7 & 302.09/147.67 & 148.22/100.56\\  
    \hline
    \end{tabular}
    \end{adjustbox}
    \caption{Cross-domain test results of HMR with the same supervision 2D\&3D KPTs. MPJPE \& PA-MPJPE are presented in unit of mm. 
    Due to limited amount of data, all released part of HuMMan are used as training data.
    }
    \label{tab:cross_results}
\end{table}

\begin{table}[h]
    \centering
    \begin{adjustbox}{max width=\linewidth}
        \begin{tabular}{c|c|c|cc}
        \toprule[1.5pt]
            Datasets & Supervision & Train & MPJPE & PA \\ \hline
            HuMMan & 2D + 3D KPTs + SMPL & FT & 85.4 & 49.2  \\ \hline
            HuMMan & 2D KPTs + SMPL & FT & 78.9 & 49.4  \\ \hline
            FreeMan & 2D + 3D KPTs+ SMPL & FT & 76.5 & \textbf{48.3}  \\ \hline
            FreeMan & 2D KPTs + SMPL & FT & \textbf{76.1} & 48.9  \\ 
        \bottomrule[1.5pt]
        \end{tabular}
    \end{adjustbox}
    \vspace{-3mm}
    \caption{Finetuning pre-trained PARE model on FreeMan and HuMMan. FreeMan can bring larger improvement compared with HuMMan.}
    \label{tab:rebuttal_pare_exp}
\end{table}

\subsection{2D-to-3D Pose Lifting}


For training data in 2D-to-3D pose lifting, we use Human3.6M data provided by VideoPose3D~\cite{DBLP:journals/corr/abs-1811-11742},
70\% of released data from HuMMan and training split of FreeMan, respectively. 
Following the original setting in HuMMan~\cite{cai2022humman}, we split released data of $100$ subjects into training and test set by subjects.
For FreeMan, we select one view from every session and down-sample the videos to $15$FPS, resulting in the frame number to be $350K$, which is similar to the amount of released part of HuMMan ($253K$) and much smaller than Human3.6M ($1500K$). 
Following~\cite{cai2022humman}, we unify the test set to be AIST++~\cite{li2021learn} in order to verify the generalization across datasets. And test set of FreeMan are used for reference.

During training, coordinates of 2D keypoints are normalized by height and width of corresponding images. 
Ground truth 3D poses are transferred into camera coordinate system and root of skeleton is placed to origin.
During test, since resolution of images are different among datasets, input 2D keypoints are normalized by resolution of test images.
Keypoints in COCO format are mapped to that in Human3.6M format following mmHuman3D~\cite{mmhuman3d}. 

All models are optimized using Adam optimizer with learning rate of $10^{-4}$ on one NVIDIA RTX-3090-24GB. 
SimpleBaseline~\cite{DBLP:journals/corr/MartinezHRL17}, VideoPose3D~\cite{DBLP:journals/corr/abs-1811-11742} are trained for 80 epochs with batch size of $1024$ and PoseFormer~\cite{DBLP:journals/corr/abs-2103-10455}, MHFormer~\cite{DBLP:journals/corr/abs-2111-12707}, PoseFormerV2~\cite{zhao2023poseformerv2} are trained for $25$ epochs with batch size of $256$ following their original settings. We show the results of VideoPose3D, PoseFormer and PoseFormerV2 in Tab.~\ref{tab:lifting supplyment}. 

\subsection{Multi-view 3D Human Pose Estimation}
In multi-view 3D human pose estimation, we use $4$ views from both Human3.6M and FreeMan as input to VoxelPose.
For Human3.6M, we follow the same processing steps as Transfusion~\cite{ma2021transfusion}. 
For FreeMan, videos from odd-indexed views in training split are downsampled by $5$ times to make data scale comparable. 
\textit{Note single frame from all input views as one group}, Human3.6M and FreeMan include $223K$ and $132K$ groups of training data, respectively.

We first finetune ResNet-50~\cite{he2016deep} backbone pre-trained on COCO with each dataset for $10$ epochs, and then optimized the latter modules in decoder for additional $15$ epochs. 
Both the two stages use Adam optimizer with a learning rate of $1e-4$ and batch size of $32$. Models are trained on $4$ NVIDIA A100-80GB GPUs.
To solve the difference between joint definitions, we select $13$ common joints between Human3.6M and COCO format, and then use the mid-points of the \textit{left \& right hips} and \textit{left \& right shoulders} to generate \textit{mid-hip} and \textit{neck}. 
In experiments, \textit{mid-hip} is used as the root joint. 
The images are all cropped by human bounding boxes and then resized to make short edges the same. 

In Tab.~\ref{tab:multiview}, we report recall and MPJPE@500mm of each experiment.
In calculation of Recall@$500$mm, only predictions with MPJPE smaller than $500$mm are treated as positive predictions and a higher recall value refers to higher successful rate to locate humans in space. And only positive predicted poses contribute to MPJPE in the final column.

Without ground truth root location, the model trained on Human3.6M is unable to locate human in cross-domain test and thus corresponding MPJPE is not available.
Even though the ground truth root positions are given, recall value and MPJPE of model trained on Human3.6M are still $96.20$\% and $103.02$mm, which is lower than that of model trained on FreeMan in cross-domain test without GT root (96.68\% \& 61.29mm), demonstrating that our training set has better transferability and test set is more challenging.

\begin{table}[t!]
\centering
\begin{adjustbox}{max width=\linewidth}
\begin{tabular}{c | c c c c}
\toprule[1.5pt]
    \multirow{1}{*}{Algorithm} &\multirow{1}{*}{Train} &
    \multirow{1}{*}{Test} &
    \multirow{1}{*}{MPJPE (mm)$\downarrow$} &\multirow{1}{*}{PA (mm)$\downarrow$} \\ \cline{1-5}
    \multirow{5}{*}{VideoPose3D} & FreeMan & FreeMan & 88.68 & 49.17 \\  
    & FreeMan$^{\dag}$ & FreeMan &73.98 & 45.22 \\  
    &Human3.6M &AIST++ &190.46 &146.98 \\ 
    &HuMMan &AIST++ &265.10 &125.56 \\ 
    & \cellcolor{gray!10}FreeMan & \cellcolor{gray!10}AIST++ & \cellcolor{gray!10}146.66 & \cellcolor{gray!10}99.01\color{Green}$\uparrow_{21.15\%}$\\ 
    &\cellcolor{gray!20}FreeMan$^{\dag}$ & \cellcolor{gray!20}AIST++ & \cellcolor{gray!20}141.84 & \cellcolor{gray!20}94.59\color{Green}$\uparrow_{24.66\%}$ \\ \hline
    
    \multirow{5}{*}{PoseFormer} & FreeMan & FreeMan & 92.94 & 64.91 \\ 
    & FreeMan$^{\dag}$ & FreeMan &77.68 & 54.39 \\ 
    &Human3.6M &AIST++ &179.54 &151.38 \\ 
    &HuMMan &AIST++ &158.13 &96.98 \\
    & \cellcolor{gray!10}FreeMan & \cellcolor{gray!10}AIST++ & \cellcolor{gray!10}133.39 & \cellcolor{gray!10}90.10\color{Green}$\uparrow_{7.09\%}$\\
    &\cellcolor{gray!20}FreeMan$^{\dag}$ &\cellcolor{gray!20}AIST++ &\cellcolor{gray!20}133.89 &\cellcolor{gray!20}84.68\color{Green}$\uparrow_{14.52\%}$ \\  \hline

    \multirow{5}{*}{PoseFormerV2} & FreeMan & FreeMan & 92.11 & 64.91 \\ 
    & FreeMan$^{\dag}$ & FreeMan & 90.81 & 55.98 \\ 
    &Human3.6M &AIST++ & 236.23 & 154.93 \\ 
    &HuMMan &AIST++ &205.73 &103.80 \\
    & \cellcolor{gray!10}FreeMan & \cellcolor{gray!10}AIST++ & \cellcolor{gray!10}131.13 & \cellcolor{gray!10}87.24\color{Green}$\uparrow_{15.95\%}$\\
    &\cellcolor{gray!20}FreeMan$^{\dag}$ &\cellcolor{gray!20}AIST++ &\cellcolor{gray!20}113.89 &\cellcolor{gray!20}80.61\color{Green}$\uparrow_{22.34\%}$ \\
    \bottomrule[1.5pt]
\end{tabular}
\end{adjustbox}
\caption{Performance of methods with different training and testing datasets in 2D-to-3D Pose Lifting. PA stands for PA-MPJPE. 
$^{\dag}$ refer to experiments with the whole training set of FreeMan. Smaller MPJPE and PA-MPJPE indicate better performance.
Highlighted rows show training on our dataset achieves the best performance in the transfer test. {\color{Green}$\uparrow$} refers to the improvement relative to HuMMan.} 
\label{tab:lifting supplyment}
\end{table}

\begin{table}[t!]
    \centering
    \begin{adjustbox}{max width=\linewidth}
    \begin{tabular}{cccc}
    \toprule[1.5pt]
         Train & Test & Recall@500mm (\%) $\uparrow$ & MPJPE (mm) $\downarrow$ \\ 
         \hline
         Human3.6M & Human3.6M  & 100 & 25.95 \\
         Human3.6M & FreeMan& 0.06 & - \\
         Human3.6M & FreeMan (w/ GT Root) & 96.20 & 154.41 \\
         \hline
         FreeMan & FreeMan   & 99.97 & 26.61 \\
         \rowcolor{gray!10}FreeMan & Human3.6M  & 96.68 & 62.37\\
         \rowcolor{gray!20}FreeMan & Human3.6M (w/ GT Root) & 100.00 & 58.30\\
         \hline
         FreeMan & FreeMan$^{\dag}$ & 99.98 & 35.04 \\
    \bottomrule[1.5pt]
    \end{tabular}
    \end{adjustbox}
    \caption{Results of VoxelPose~\cite{voxelpose} for Multi-View 3D Pose Estimation. Recall@500mm refer to ratio of predictions with MPJPE smaller than $500$mm, MPJPE here has not threshold for all keypoints.
    FreeMan$^{\dag}$ represents test set of even indexed cameras. Ground truth root position (GT Root) is not used if not specified. 
    Rows highlighted shows the best setting in cross-domain test.}
    \label{tab:multiview}
\end{table}

\subsection{Neural Rendering of Human Subjects}
\subsubsection{Implementation Details}
We use $128$ samples per ray and train for $400K$ iterations with the Adam optimizer as the setting in \cite{weng2022humannerf}. Samely, to improve the quality of our results, we have increased the number of rays sampled for the foreground subject, as identified by the segmentation masks. We achieve this by implementing a random ray sampling method that assigns a higher probability of $0.8$ to foreground subject pixels and a lower probability of $0.2$ to the background region. The resize scale of the image is set to $0.5$. It takes about $48$ hours to train on one NVIDIA RTX-3090-24GB for each one.

In order to ensure the quality of training, the number of frames of video clips in different scenes is in the interval of $300$ to $1200$ frames. The selected ten clips contain a variety of actions, ranging from slow and deliberate movements (such as warm-up exercises) to fast and energetic ones (such as dancing).

\subsubsection{Visualization Results}
\begin{figure}
    \centering
    \includegraphics[width=\linewidth]{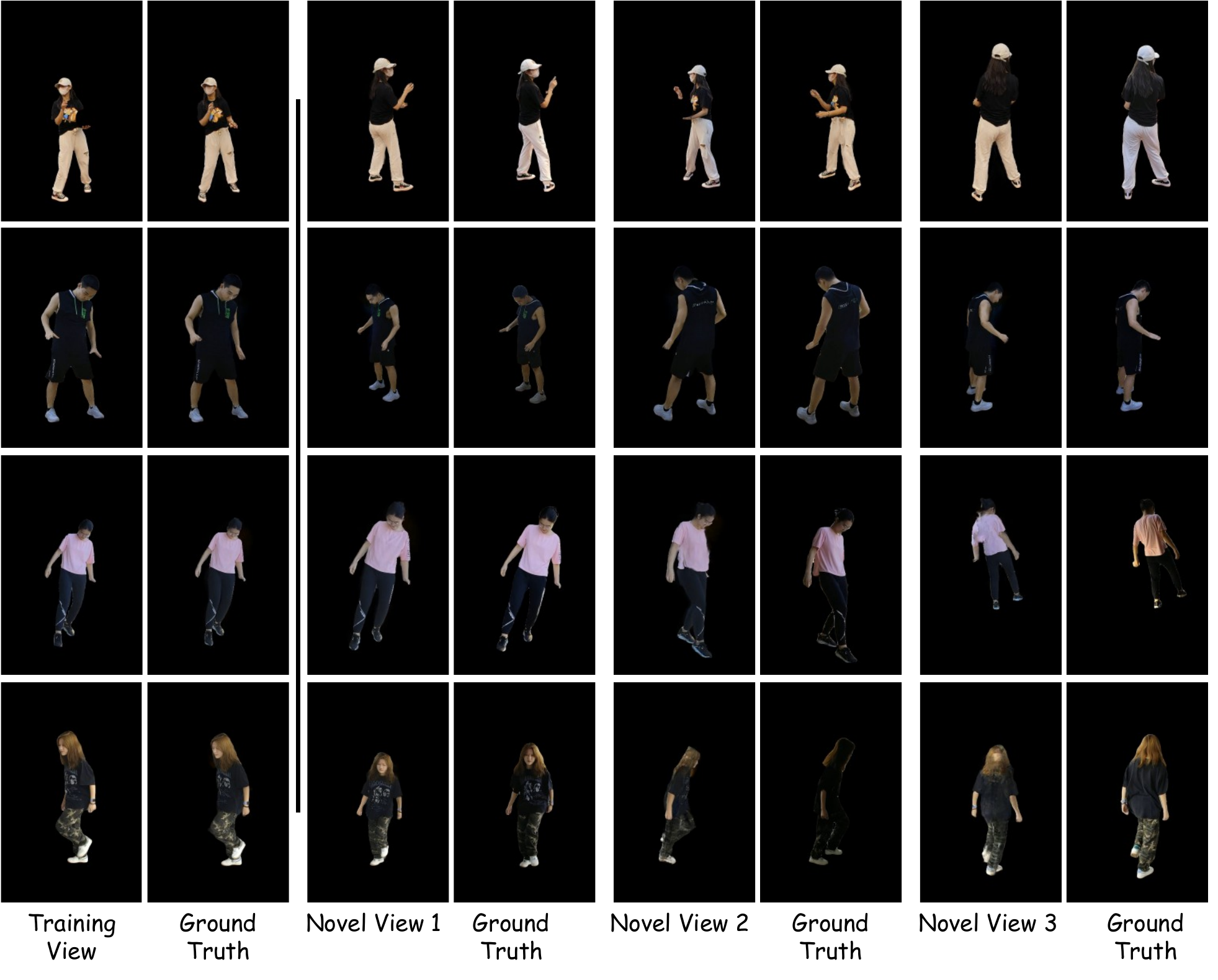}
    \caption{Rendering results for selected sessions of $30$FPS. The lower two rows present cases of bad rendering results.}
    \label{fig:render}
\end{figure}

We show the visualization results of the reconstruction of the selected videos in FreeMan dataset at Fig.~\ref{fig:render}. The above two lines reflect the results of relatively good reconstruction, while the following two lines reflect the results of relatively poor reconstruction. This indicates that FreeMan has sufficient diversity and challenges for human reconstruction.

\subsubsection{Experiments on data of 60FPS}
\begin{figure}
    \centering
    \includegraphics[width=\linewidth]{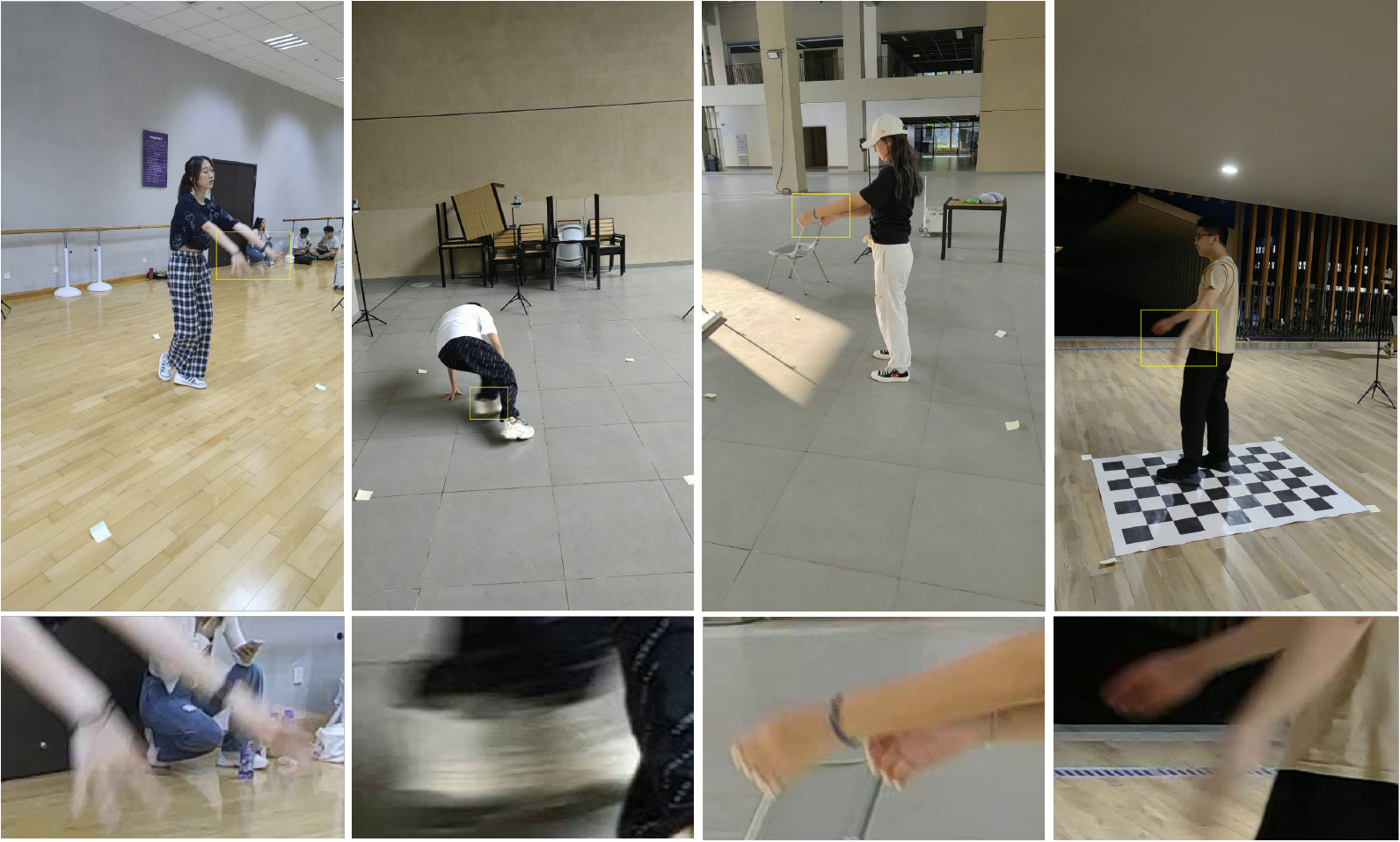}
    \caption{Example images of motion blur on human body in sessions of $30$FPS. Image patches within bounding boxes at upper row are shown at lower row.}
    \label{fig:blur}
\end{figure}

Due to the occurrence of blur as Fig.~\ref{fig:blur} in body parts such as hands and feet when moving at high speeds, we collect videos at $60$FPS to provide higher quality ground truth. We conduct experiments on two video clips from Park and Square scenes, and the experimental results are as Tab.~\ref{humannerf_60}. The results indicate that FreeMan remains highly challenging for human neural rendering in natural lighting conditions.

\begin{table}[h]
\centering
\begin{tabular}{c|ccc}
\toprule[1.5pt]
     Scene & PSNR$\uparrow$ & SSIM$\uparrow$ & LPIPS$^{*}\downarrow$ \\ 
     \hline
     Square & $23.99$ & $0.9389$ & $88.10$  \\
     Park & $24.43$ & $0.9527$ & $61.84$ \\
\bottomrule[1.5pt]
\end{tabular}
\caption{Neural rendering in $60$FPS results by using HumanNeRF~\cite{weng2022humannerf}. \texttt{LPIPS$^{*}$} = \texttt{LPIPS} $\times 10^3$.}
\label{humannerf_60}
\end{table}

\begin{figure*}[th]
    \centering
        \includegraphics[width=0.9\linewidth]{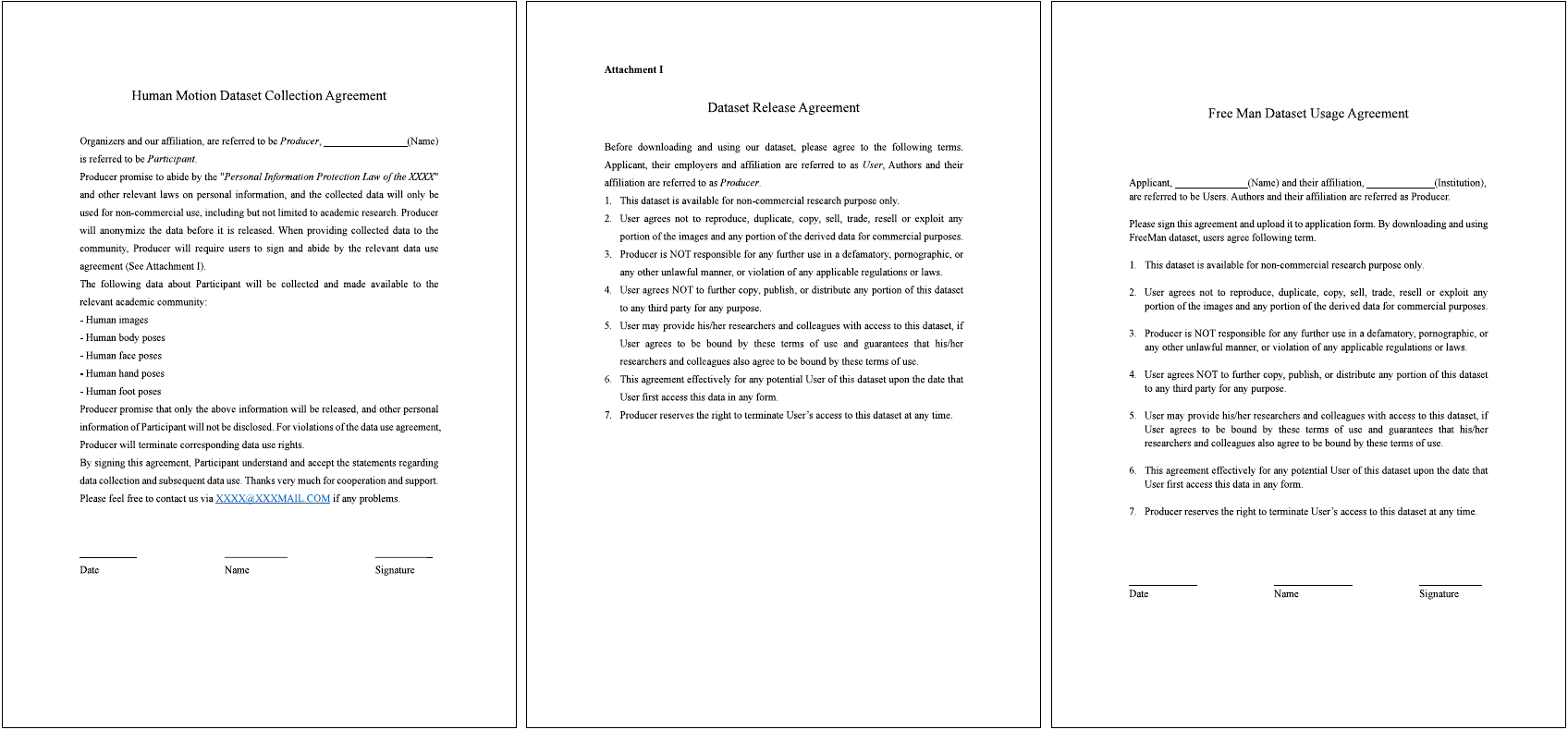}
    \caption{\textbf{Left}: Data collection agreements for actors involved in FreeMan. \textbf{Middle}: Illustration of users' responsibility and usage agreement for actors. \textbf{Right}: Usage agreement for dataset users. Information collected will not be published are only for backup. Identity information are omitted.}
    \label{fig:supp_use_agreement}
\end{figure*}

\section{Dataset Documentations}
\label{sec:freeman_usage}
FreeMan is available for academic communities to boost related researches.
Example code to load FreeMan in pytorch is available on Github and data storage structure are illustrated in this section.

\subsection{Data Format}
FreeMan consists of multi-view human motion data and corresponding 2D / 3D human pose annotations. 
All data are separated into videos, camera parameters, bounding boxes and keypoints annotations basd on data type. 
For each session, human motion videos of all views are stored in format of $mp4$ and there are $8$ synchronized videos for $8$ views. Camera parameters, including image resolution, camera intrinsic parameters and camera extrinsic parameters, are saved in $JSON$ format.

Human keypoint annotations are encoded into format of $npy$, which is also known as numpy array. 2D poses of one session are stored with an array whose shape is $[V, F, J, 2]$, where $V$ for view indexes, $F$ for frame number, $J$ for total number of joints and keypoint locations are given by $(x, y)$ coordinates in unit of pixels.
3D poses are stored in an array with shape of $[F, J, 3]$, and 3D keypoint locations are provided by $(x, y, z)$ in world coordinate system.

\subsection{License \& Ethical Impact}

As FreeMan is constructed for research purpose only, FreeMan adopts license of CC BY-NC 4.0 (Non-Commercial use only).
Furthermore, subsequent users who are granted access to the dataset are required to sign relevant usage agreements and provide backup information. 
This is done to safeguard the privacy and security of individuals associated with the FreeMan dataset and prevent data misuse.

All actors involved in FreeMan are recruited on basis of voluntary and well informed of data collection purpose.
All volunteers signed a data collection agreement which declares project proposal and data to be released, which is shown in Fig.~\ref{fig:supp_use_agreement}.
To protect the privacy of all participants, we anonymized all data in the dataset by removing any personally identifiable information. 

Major potential social impacts for human related research are about privacy leakage.
We have taken several steps to protect the privacy and anonymity of the individuals involved in the dataset. 
All data subjects provided informed consent before participating, and we have anonymized all data to the best of our abilities to ensure the non-disclosure of any personal information of the actors.
Furthermore, we understand the importance of data governance and the need to address potential misuse or unintended consequences. 
We encourage researchers and users of our dataset to handle the data responsibly, following ethical guidelines and respecting privacy considerations. 
Before getting access to our data, researchers will be required to sign an agreement to obey our license. 
We are committed to ongoing discussions and collaborations with experts in the field to address any concerns and ensure that our work contributes positively to the research community while minimizing any potential negative social impact.

\subsection{Maintenance Plan}

To access FreeMan data, users are required to sign a dataset usage agreement that illustrates responsibilities and requirements.
After submitting signed agreement and basic information via online forms, they can download FreeMan from dataset host. 
All users should abide the relevant data use agreement and use rights will be terminated for any violations of data use agreement.
Application procedure requires applicants to submit their basic information for backup and our data are available on \href{https://huggingface.co/datasets/wjwow/FreeMan}{Huggingface} and \href{https://opendatalab.com/wangjiongwow/FreeMan}{OpenDataLab} for research community.

\subsection{Agreements}
Specifically, we present agreements for both actors and users in Fig.~\ref{fig:supp_use_agreement}. 
The leftmost two pages are for actors in the project. 
The first page is to explain this project and show data type while the second page is to show how our data will be publiced and used.

The last page is required for users to sign before access FreeMan, explaining users' responsibility. Information is collected for backup purposes only.

\end{document}